\documentclass[a4paper,11pt]{article}

\usepackage[T1]{fontenc}
\usepackage[utf8x]{inputenc}
\usepackage[english]{babel}
\usepackage[colorlinks=true, allcolors=blue]{hyperref}
\usepackage{comment}
\usepackage{todonotes}
\usepackage{longtable}

\newcommand{\email}[1]{\href{mailhttps://pt.overleaf.com/project/616c5d22237b5c589724ddb7to:#1}{\tt{\nolinkurl{#1}}}}
\newcommand{\orcid}[1]{ORChttps://pt.overleaf.com/project/616c5d22237b5c589724ddb7ID: \href{https://orcid.org/#1}{\tt{\nolinkurl{#1}}}}

\usepackage[sfdefault,lf]{carlito}
\usepackage[parfill]{parskip}

\usepackage{fancyhdr}
\usepackage{natbib}
\usepackage{authblk}
\usepackage[a4paper,top=3cm,bottom=2cm,left=3cm,right=3cm,marginparwidth=1.75cm]{geometry}

\usepackage{amsmath, amssymb}
\usepackage{graphicx}
\usepackage{booktabs}
\usepackage{tikz}
\usetikzlibrary{bayesnet}
\usetikzlibrary{arrows}
\usetikzlibrary{backgrounds}
\usepackage{float}

\title{Identifying latent disease factors differently expressed in patient subgroups using group factor analysis}
\author[1,2]{Fabio S. Ferreira}
\author[3]{John Ashburner}
\author[4]{Arabella Bouzigues}
\author[5]{Chatrin Suksasilp}
\author[4]{Lucy L. Russell}
\author[4]{Phoebe H. Foster}
\author[4]{Eve Ferry-Bolder}
\author[6]{John C. van Swieten}
\author[6]{Lize C. Jiskoot}
\author[6]{Harro Seelaar}
\author[7]{Raquel Sanchez-Valle}
\author[8]{Robert Laforce}
\author[9]{Caroline Graff}
\author[10]{Daniela Galimberti}
\author[11]{Rik Vandenberghe}
\author[12]{Alexandre de Mendon\c{c}a}
\author[13]{Pietro Tiraboschi}
\author[14]{Isabel Santana}
\author[15]{Alexander Gerhard}
\author[16]{Johannes Levin}
\author[17]{Sandro Sorbi}
\author[18]{Markus Otto}
\author[19]{Florence Pasquier}
\author[20]{Simon Ducharme}
\author[21]{Chris R. Butler}
\author[22]{Isabelle Le Ber}
\author[23]{Elizabeth Finger}
\author[24]{Maria Carmela Tartaglia}
\author[25]{Mario Masellis}
\author[26]{James B. Rowe}
\author[27]{Matthis Synofzik}
\author[28]{Fermin Moreno}
\author[29]{Barbara Borroni}
\author[30,**]{Samuel Kaski}
\author[4,**]{Jonathan D. Rohrer}
\author[1,2,*,**]{Janaina Mour\~{a}o-Miranda}
\author[***]{the Genetic FTD Initiative (GENFI)}
\affil[1]{\small Centre for Medical Image Computing, Department of Computer Science, University College London, London, UK}
\affil[2]{\small Max Planck University College London Centre for Computational Psychiatry and Ageing Research, University College London, UK}
\affil[3]{\small Wellcome Centre for Human Neuroimaging, University College London, London, UK}
\affil[4]{\small Dementia Research Centre, Department of Neurodegenerative Disease, UCL Queen Square Institute of Neurology, London, UK}
\affil[5]{\small Institute of Cognitive Neuroscience, University College London, London, UK}
\affil[6]{\small Department of Neurology, Erasmus Medical Centre, Rotterdam, Netherlands}
\affil[7]{\small Alzheimer's disease and Other Cognitive Disorders Unit, Neurology Service, Hospital Clinic, Institut d'Investigacions Biomediques August Pi I Sunyer, University of Barcelona, Barcelona, Spain}
\affil[8]{\small Clinique Interdisciplinaire de Memoire, Departement des Sciences Neurologiques, CHU de Quebec, and Faculte de Medecine, Universite Laval, QC, Canada}
\affil[9]{\small Department of Neurobiology, Care Sciences and Society; Center for Alzheimer Research, Division of Neurogeriatrics, Bioclinicum, Karolinska Institutet, Solna, Sweden}
\affil[10]{\small Fondazione Ca Granda, IRCCS Ospedale Policlinico, Milan, Italy}
\affil[11]{\small Laboratory for Cognitive Neurology, Department of Neurosciences, KU Leuven, Leuven, Belgium}
\affil[12]{\small Faculty of Medicine, University of Lisbon, Lisbon, Portugal.}
\affil[13]{\small Fondazione IRCCS Istituto Neurologico Carlo Besta, Milano, Italy}
\affil[14]{\small University Hospital of Coimbra (HUC), Neurology Service, Faculty of Medicine, University of Coimbra, Coimbra, Portugal}
\affil[15]{\small Division of Psychology Communication and Human Neuroscience, Wolfson Molecular Imaging Centre, University of Manchester, Manchester, UK}
\affil[16]{\small Department of Neurology, Ludwig-Maximilians Universitat Munchen, Munich, Germany}
\affil[17]{\small Department of Neurofarba, University of Florence, Italy}
\affil[18]{\small Department of Neurology, University of Ulm, Germany}
\affil[19]{\small Univ Lille, France}
\affil[20]{\small Douglas Mental Health University Institute, Department of Psychiatry, McGill University, Montreal Canada}
\affil[21]{\small Nuffield Department of Clinical Neurosciences, Medical Sciences Division, University of Oxford, Oxford, UK}
\affil[22]{\small Sorbonne Universite, Paris Brain Institute $-$Institut du Cerveau $-$ ICM, Inserm U1127, CNRS UMR 7225, AP-HP $-$ Hopital Pitie-Salpetriere, Paris, France}
\affil[23]{\small Department of Clinical Neurological Sciences, University of Western Ontario, London, ON, Canada} 
\affil[24]{\small Tanz Centre for Research in Neurodegenerative Diseases, University of Toronto, Toronto, ON, Canada}
\affil[25]{\small Sunnybrook Health Sciences Centre, Sunnybrook Research Institute, University of Toronto, Toronto, Canada}
\affil[26]{\small Department of Clinical Neurosciences and Cambridge University Hospitals NHS Trust, University of Cambridge, UK}
\affil[27]{\small Department of Neurodegenerative Diseases, Hertie-Institute for Clinical Brain Research and Center of Neurology, University of Tubingen, Tubingen, Germany}
\affil[28]{\small Cognitive Disorders Unit, Department of Neurology, Hospital Universitario Donostia, 20014, San Sebastian, Spain}
\affil[29]{\small Neurology Unit, Department of Clinical and Experimental Sciences, University of Brescia, Brescia, Italy}
\affil[30]{\small Helsinki Institute for Information Technology HIIT, Department of Computer Science, Aalto University, Finland}
\affil[*]{\small Corresponding author: \email{j.mourao-miranda@ucl.ac.uk}}
\affil[**]{These authors contributed equally to this work.}
\affil[***]{A list of collaborators of the GENFI consortium is provided in the Supplementary Material}
\date{}

\begin{document}
\maketitle
\thispagestyle{fancy}

\begin{abstract}
In this study, we propose a novel approach to uncover subgroup-specific and subgroup-common latent factors addressing the challenges posed by the heterogeneity of neurological and mental disorders, which hinder disease understanding, treatment development, and outcome prediction. The proposed approach, sparse Group Factor Analysis (GFA) with regularised horseshoe priors, was implemented with probabilistic programming and can uncover associations (or latent factors) among multiple data modalities differentially expressed in sample subgroups. Synthetic data experiments showed the robustness of our sparse GFA by correctly inferring latent factors and model parameters. When applied to the Genetic Frontotemporal Dementia Initiative (GENFI) dataset, which comprises patients with frontotemporal dementia (FTD) with genetically defined subgroups, the sparse GFA identified latent disease factors differentially expressed across the subgroups, distinguishing between "subgroup-specific" latent factors within homogeneous groups and "subgroup common" latent factors shared across subgroups. The latent disease factors captured associations between brain structure and non-imaging variables (i.e., questionnaires assessing behaviour and disease severity, neuropsychological tasks and medical assessments of disease severity) across the different genetic subgroups, offering insights into disease profiles. Importantly, two latent factors were more pronounced in the two more homogeneous FTD patient subgroups (progranulin (\textit{GRN}) and microtubule-associated protein tau (\textit{MAPT}) mutation), showcasing the method's ability to reveal subgroup-specific characteristics. These findings underscore the potential of sparse GFA for integrating multiple data modalities and identifying interpretable latent disease factors that can improve the characterization and stratification of patients with neurological and mental health disorders.
\end{abstract}

\section{Introduction}
The heterogeneity of neurological and mental health disorders has been a key confound to disease understanding, treatment development and outcome prediction, as patient populations are thought to include multiple disease pathways that selectively respond to treatment \citep{Kapur2012}. These challenges are reflected in poor treatment outcomes; for instance, in depression, approximately only 40\% of patients remit after first-line antidepressant treatment or psychotherapy \citep{Amick2015, Cuijpers2014, Fava1996, Trivedi2006}.

Diagnostic categories in psychiatry have historically been defined based on signs and symptoms, prioritising diagnostic agreement between clinicians, rather than underlying biological mechanisms \citep{Freedman2013, Robins1970}. Resultingly, the usefulness of supervised machine learning methods as diagnostic tools for mental health disorders (i.e., classifying patients vs. healthy controls) is questionable, as they may simply inherit the flaws of current diagnostic categories. Additional challenges in neurological and mental health disorders are comorbidity (i.e., individuals with one disorder often develop another disorder during their lifespan) and that different disorders can share similar symptoms \citep{Kessler2005}. To address the limitations of current diagnostic categories in psychiatry, the National Institute of Mental Health launched the Research Domain Criteria framework (RDoC) in 2009 (\url{https://www.nimh.nih.gov/research/research-funded-by-nimh/rdoc}) as an attempt to move beyond diagnostic categories and ground psychiatry within neurobiological constructs that combine multiple levels of measures or sources of information \citep{Insel2010}.

Multivariate methods, such as Canonical Correlation Analysis (CCA) and related methods, that do not rely on the diagnostic categories, have been widely used to uncover latent disease dimensions capturing associations between brain imaging and non-imaging data (e.g., self-report questionnaires, cognitive tests and genetics). The identified latent dimensions provide information on how a set of non-imaging features (e.g. symptoms and behaviours) jointly map into brain features. This approach can ground neurological and mental health disorders within neurobiology and provide a better understanding of the disorders. For instance, CCA has been applied jointly with principal component analysis to investigate associations between brain connectivity, demographics and behaviour in healthy population \citep{Smith2015, Bijsterbosch2018, Li2019}, healthy and clinically depressed adolescents and young adults \citep{Mihalik2019} and children \citep{Alnaes2020}. 

Moreover, sparse variants of CCA  have been extensively applied to identify associations between genomics and brain imaging \citep{Waaijenborg2008, Witten2009a, Witten2009b, Parkhomenko2009}. The most commonly used sparse CCA approach \citep{Witten2009a} assumes that the within-modality covariance matrices in the CCA optimization problem are identity matrices to enable a simple solution,  so the model becomes equivalent to a sparse Partial Least Square (PLS) model (see Supplementary Material in \citet{Mihalik2022} for a detailed explanation). This sparse PLS approach has been used in many studies to identify latent dimensions of association between different imaging modalities and between imaging and non-imaging features. \citet{Avants2010} applied sparse PLS to capture associations between brain structure and diffusion tensor imaging in Alzheimer's disease and frontotemporal dementia patients. \citet{Monteiro2016} used sparse PLS to uncover associations between brain structure and demographic and clinical/cognitive data in a sample of healthy controls and patients with Alzheimer's disease and mild cognitive impairment. More recently, sparse PLS has been applied to find associations between behaviour, clinical, and multimodal imaging phenotypes in psychosis \citep{Moser2018}, and between functional connectivity and psychiatric symptoms in a large sample of young people \citep{Xia2018}. \citet{Mihalik2020a} used sparse PLS to identify associations between brain structure, demographics and behaviour in a sample of healthy and depressed adolescents and young adults. \citet{Adams2024} applied sparse PLS to identify latent dimensions of association between brain structure and psychosocial variables in the Adolescent Brain and Cognitive Development (ABCD) cohort.

\begin{figure}[H]
\centering
\includegraphics[width=\linewidth]{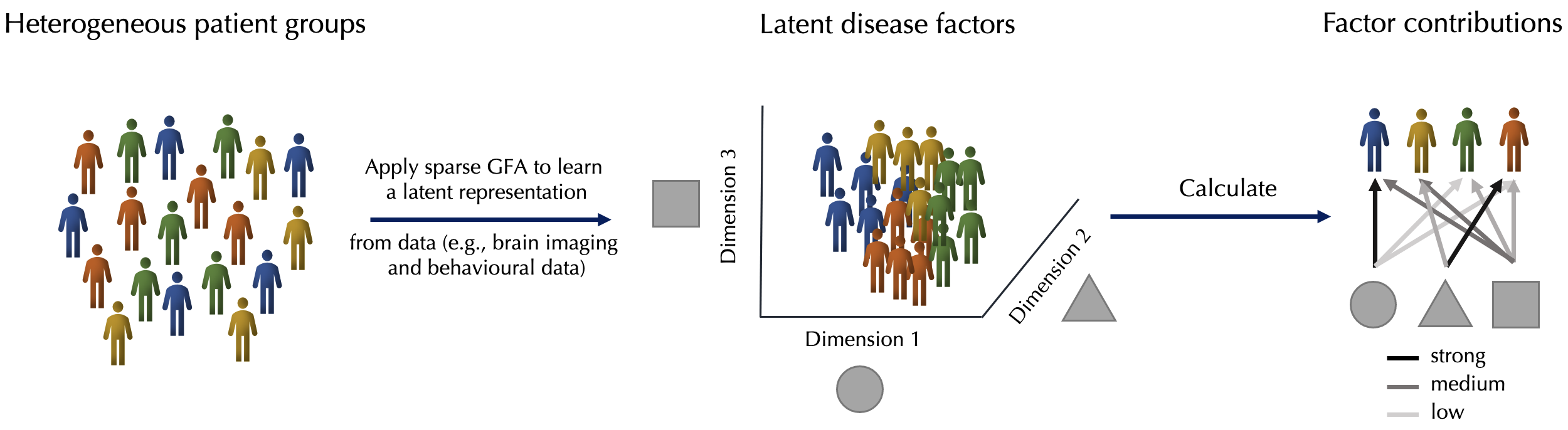}
\caption{Dimensional approach for subgroup characterisation: data-derived latent disease factors can help disentangle the heterogeneity of neurological and mental health disorders. Latent variable models, such as sparse GFA,  can identify latent dimensions from patients' data (i.e., latent disease factors) that can be differently expressed (i.e., different factor contributions) across different subgroups of patients.}
\label{fig:latent_factors}            
\end{figure}

Despite the success that these methods have shown in recent years, they have some limitations: 1) CCA and equivalent methods do not provide an inherently robust approach to infer the relevant associations - this is usually done by assessing the statistical significance of the associations using permutation tests on the whole data set \citep{Smith2015, Mihalik2019, Winkler2020} or on hold-out sets \citep{Monteiro2016, Mihalik2020a}; 2) the associations within subsets of data modalities, which might explain important variance in the data, are not modelled; 3) it is not straightforward to integrate more than two modalities in the analysis (although there are approaches for applying CCA for more than two modalities \citep{Qi2018, Zhuang2020}); 4) more importantly, CCA and equivalent methods find associations that are present in the whole sample and therefore might not identify associations expressed differentially in subgroups of patients, which can help understanding the heterogeneity of neurological and psychiatric disorders (Fig. \ref{fig:latent_factors}). 

Group Factor Analysis (GFA) \citep{Virtanen2012, Klami2015} has been proposed as a generalization of factor analysis (Thurstone et al. 1931) that aims to investigate associations (termed latent factors) among more than two data modalities. GFA is a generative probabilistic model that uses Bayesian inference and decomposes the multiple data modalities (termed groups) between across-modalities and within-modalities associations, while controlling for model complexity, i.e., selecting the optimal number of factors. Previously, we have shown that GFA can handle missing data and scales well to high dimensional data by applying it to data from the Human Connectome Project (HCP) \citep{Essen2013} to uncover associations between brain connectivity and non-imaging features (e.g., demographics, psychometrics and other behavioural features) \citep{Ferreira2022}. The results were in line with previous literature \citep{Smith2015} and showed consistency across different experiments with complete and incomplete data (which included missing values). Nonetheless, GFA does not allow feature-wise sparsity, which is particularly useful for feature selection and model interpretation, nor sample-wise sparsity, which is important for identifying latent factors differently expressed across the sample. 

\citet{Bunte2016} proposed a sparse extension of GFA by adding Bayesian shrinkage priors (e.g., spike-and-slab priors) to the GFA model to impose sparsity over samples and features, respectively. The spike-and-slab priors have been widely used in sparse Bayesian models and have shown good performance in practice \citep{Piironen2017, VanErp2019}. However, an important limitation of this model is the fact that due to the discrete nature of the spike-and-slab prior, model inference can be quite slow making it difficult to apply it to multiple high-dimensional data, e.g. neuroimaging and genetic data. Alternative continuous shrinkage priors, such as the horseshoe prior \citep{Carvalho2009}, have been proposed to provide more efficient inference by using inference methods such as Hamiltonian Monte Carlo (HMC) \citep{Neal2011, Betancourt2018} alongside probabilistic programming packages/libraries (such as Stan \citep{Stan2019} and PyMC \citep{Salvatier2016}), while obtaining similar performance in practice \citep{Piironen2017, VanErp2019}.

In this study, we propose a new sparse GFA method by replacing the spike-and-slab priors of Bunte et al.'s model with regularised horseshoe priors \citep{Piironen2017}, implemented with probabilistic programming software. We show that the proposed model can uncover sparse associations among multiple data modalities and identify latent disease factors differently expressed in subgroups of patients. Our implementation of sparse GFA was applied to synthetic data and the Genetic Frontotemporal Dementia Initiative (GENFI) dataset, which includes patients with genetic frontotemporal dementia (FTD). A large proportion of FTD cases are caused by mutations in progranulin (\textit{GRN}), microtubule-associated protein tau (\textit{MAPT}) and chromosome 9 open reading frame 72 (\textit{C9orf72}) \citep{Snowden2012, Koskinen2013}. \textit{GRN} and \textit{MAPT} mutations are associated with distinct phenotypes representing more homogeneous groups, whereas \textit{C9orf72} is known to be a heterogeneous group \citep{Mahoney2012}. The GENFI dataset serves as a real data set, with known subgroups, where we can test the hypothesis that sparse GFA can identify latent disease factors that are differently expressed across subgroups of patients. More specifically, we hypothesise that sparse GFA will identify latent disease factors that are more strongly expressed within the homogeneous subgroups ("subgroup-specific" latent factors) as well as latent factors that are similarly expressed across the different subgroups ("subgroup-common" latent factors). Each patient subgroup can then be characterized by a combination of "subgroup-specific" and "subgroup-common" latent factors expressed with different strengths as represented in (Fig. \ref{fig:latent_factors}). This approach enables using a dimensional perspective to characterise patient subgroups, in contrast with a categorical approach that conceptualises patient subgroups as different clusters.

\section{Methods}
Here, we start by briefly describing Group Factor Analysis (Section \ref{Methods_GFA}), which is followed by descriptions of our new implementation of sparse GFA using horseshoe priors (Section \ref{Methods_sGFA_hsprior}). We then describe the model inference and implementation (Section \ref{Methods_MI}). These sections are followed by a brief description of the synthetic data (Section \ref{Synthetic_data}) and the GENFI dataset (Section \ref{GENFI_data}) used in this study. We end this section by describing the approach used to assess the robustness of the inferred factors (Section \ref{Methods_robcomps}).

\subsection{Group Factor Analysis}
\label{Methods_GFA}
\begin{figure}[H]
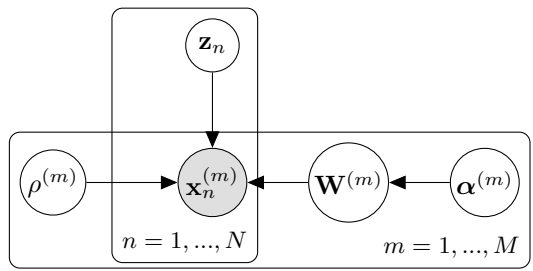

\centering
    \tikz{  
            \node[obs] (x) {$\mathbf{x}_{n}^{(m)}$};%
            \node[latent,above= of x] (z) {$\mathbf{z}_{n}$}; %
            \node[latent,left= 1.2 of x] (t) {$\rho^{(m)}$}; %
            \node[latent,right= 0.8 of x] (w) {$\mathbf{W}^{(m)}$}; %
            \node[latent,right= 0.8 of w] (al) {$\boldsymbol{\alpha}^{(m)}$}; %
            \plate{plate1} {(x)(z)} {$n = 1,...,N$}; %
            \plate{plate2} {(t)(x)(w)(al)} {$m = 1,...,M$}; %
            \edge {z} {x}
            \edge {w} {x}
            \edge {t} {x}
            \edge {al} {w}
        }
\caption{Graphical representation of GFA.}
\label{fig:GFA_graph}            
\end{figure}

In the GFA problem, we assume that $N$ observations, stored in $\mathbf{X} \in \mathbb{R}^{D \times N}$, have disjoint $M$ partitions of features $D_{m}$ called groups ($\mathbf{X}^{(m)} \in \mathbb{R}^{D_{m} \times N}$ for the $m$th group, such that $D = \sum_{m=1}^M D_m$), which can be interpreted as different data modalities. GFA finds a set of $K$ factors (i.e, columns of the loading matrix $\mathbf{W} \in \mathbb{R}^{D \times K}$) that describe $\mathbf{X}$ so that the associations between data modalities can be separated from those within data modalities. This can be solved by formulating a joint factor model for $\mathbf{X}$ (Fig. \ref{fig:GFA_graph}), where each data modality is generated as follows \citep{Virtanen2012, Klami2015}:
\begin{equation} \label{eq1}
\begin{gathered}
    \mathbf{z}_{n} \sim \mathcal{N}(\mathbf{0}, \mathbf{I}_{K}), \\
    \mathbf{x}^{(m)}_{n} \sim \mathcal{N}(\mathbf{W}^{(m)}\mathbf{z}_{n}, \ \rho^{(m)^{-1}} \mathbf{I}),
\end{gathered}
\end{equation}
where $\rho^{(m)}$ represents the noise precision, i.e., inverse noise variance, of the $m$th modality, $\mathbf{W}^{(m)} \in \mathbb{R}^{D_{m} \times K}$ is the loading matrix of the $m$th view and $\mathbf{z}_{n} \in\mathbb{R}^{K \times 1}$ is the latent variable of $\mathbf{x}^{(m)}_n$. The model assumes zero-mean data without loss of generality. 

The  structure of the loading matrices ($\mathbf{W}$) and the corresponding latent structure (represented by $\mathbf{Z}$) is learned by imposing structured sparsity over the factors, which is achieved by adding independent Automatic Relevance Determination (ARD) priors over $\mathbf{W}$ \citep{Virtanen2012, Klami2015}:
\begin{equation} \label{eq2}
    p(\mathbf{W}|\boldsymbol{\alpha}) = \prod_{m=1}^{M} \prod_{j=1}^{D_{m}} \prod_{k=1}^{K} \mathcal{N}(w^{(m)}_{j,k}|0, (\alpha_{k}^{(m)})^{-1}), \ p(\boldsymbol{\alpha}) = \prod_{m=1}^{M} \prod_{k=1}^{K} \Gamma(\alpha^{(m)}_{k}| a_{\boldsymbol{\alpha}^{(m)}},b_{\boldsymbol{\alpha}^{(m)}})
\end{equation}

A separate ARD prior is used for each $\mathbf{W}^{(m)}$, which are chosen to be uninformative to enable the pruning of irrelevant factors. $\Gamma(\cdot)$ represents a gamma distribution with shape parameter $a_{\boldsymbol{\alpha}^{(m)}}$ and rate parameter $b_{\boldsymbol{\alpha}^{(m)}}$. These independent priors cause groups of features to be pushed close to zero for some factors $k$ ($\mathbf{w}_{k}^{(m)} \to 0$) by driving the corresponding $\alpha_{k}^{(m)}$ towards infinity. If the loadings of a factor are pushed towards zero for all data modalities, the underlying factor is pruned out. Finally, the prior distributions over the noise and latent variables $\mathbf{Z}$ are given by:
\begin{equation} \label{eq3}
    p(\boldsymbol{\rho}) = \prod_{m=1}^{M} \Gamma(\rho^{(m)}|a_{\rho^{(m)}},b_{\rho^{(m)}}) , \qquad p(\mathbf{Z}) = \prod_{k=1}^{K} \prod_{n=1}^{N} \mathcal{N}(z_{k,n}|0,1)
\end{equation}
where $\Gamma(\cdot)$ represents a gamma distribution with shape parameter $a_{\boldsymbol{\rho}^{(m)}}$ and rate parameter $b_{\rho^{(m)}}$. The hyperparameters $a_{\boldsymbol{\alpha}^{(m)}}, b_{\boldsymbol{\alpha}^{(m)}}, a_{\rho^{(m)}}, b_{\rho^{(m)}}$ can be set to a very small number (e.g., $10^{-14}$), resulting in uninformative priors. The joint distribution $p(\mathbf{X,Z,W}, \boldsymbol{\alpha}, \boldsymbol{\rho})$ is hence given by:
\begin{equation} \label{eq4}
    p(\mathbf{X,Z,W},\boldsymbol{\alpha,\rho}) = p(\mathbf{X}|\mathbf{Z,W},\boldsymbol{\rho})p(\mathbf{Z})p(\mathbf{W}|\boldsymbol{\alpha}) p(\boldsymbol{\alpha})p(\boldsymbol{\rho})
\end{equation} 

\sloppy For learning the GFA model, one needs to infer the model parameters and latent variables from data, which can be done by estimating the posterior distribution $p(\mathbf{Z},\mathbf{W}, \boldsymbol{\alpha}, \boldsymbol{\rho} | \mathbf{X})$ and marginalising out uninteresting variables. However, these marginalisations are often analytically intractable, so the posterior distribution needs to be approximated using, for instance, mean field variational approximation. For more details regarding the GFA inference, see \citet{Virtanen2012, Klami2015}.

\subsection{Sparse GFA using spike-and-slab priors}
\label{Methods_sGFA_spike}
\citet{Bunte2016} proposed a sparse extension of GFA to find biclusters, which enables feature-wise and sample-wise sparsity. Biclusters are defined as sets of rows that are similar to sets of columns in a data matrix, and vice versa. For instance, biclusters can be interpreted as subsets of individuals sharing associations among subsets of features of multiple data modalities. The biclusters are inferred by adding shrinkage priors (e.g., spike-and-slab priors) over the loading matrices and latent variables to impose sparsity over samples and features, respectively. 

As described in the previous section, GFA provides view-wise sparsity; however, in some applications, only a few subsets of features within each view might be associated with features in other views. Therefore, feature-wise sparsity is important to improve the interpretability of the model. This can be achieved by adding spike-and-slab priors \citep{Mitchell1988}, over the loading matrices \citep{Khan2014, Bunte2016}:
\begin{equation} \label{eq5}
\begin{gathered}
	w_{j,k}^{(m)}|h_{w_{j,k}}^{(m)}, \alpha_{w_{k}}^{(m)} \sim h_{w_{j,k}}^{(m)} \mathcal{N} \big(0, (\alpha_{w_{k}}^{(m)})^{-1}\big) + (1 - h_{w_{j,k}}^{m}) \delta_{0}, \\
	h_{w_{j,k}}^{(m)}|\pi_{w_{k}}^{(m)} \sim \mathrm{Bernoulli}(\pi_{w_{k}}^{(m)}), \quad \pi_{w_{k}}^{(m)} \sim \mathrm{Beta}(a_{\pi}, b_{\pi}), \quad \alpha_{w_{k}}^{(m)} \sim \Gamma(a_{\alpha_{w}},b_{\alpha_{w}})
\end{gathered}
\end{equation}   
where $h_{w_{j,k}}^{(m)}$ is binary and determines whether the component $k$ is active in the $j$-th feature of $\mathbf{X}^{(m)}$, $\pi_{w_{k}}^{(m)}$ represents the probability of $h_{w_{j,k}}^{(m)} = 1$ and $\delta_{0}$ is a delta function at $0$. $\alpha_{w_{k}}^{(m)}$ is sampled from a Gamma prior and determines the scale of the component $k$ in view $m$. If the Gamma prior is uninformative, as explained in Section \ref{Methods_GFA}, it implements view-wise sparsity. In this way, the model imposes sparsity over the views and features simultaneously, which enables finding associations between subsets of features among multiple data modalities, for instance.

Another spike-and-slab prior is added over the latent variables to impose sparsity over the samples. This is relevant if one assumes that associations are present only in the data's subsamples (e.g., subgroups of patients). The prior over the latent variables is defined as follows \citep{Bunte2016}:
\begin{equation} \label{eq6}
\begin{gathered}
	z_{n,k}|h_{z_{n,k}}, \alpha_{z_{k}} \sim h_{z_{n,k}} \ \mathcal{N} \big(0, (\alpha_{z_{k}})^{-1}\big) + (1 - h_{z_{n,k}}) \delta_{0}, \\
	h_{z_{n,k}}|\pi_{z_{k}} \sim \mathrm{Bernoulli}(\pi_{z_{k}}), \quad \pi_{z_{k}} \sim \mathrm{Beta}(a_{\pi}, b_{\pi}), \quad \alpha_{z_{k}} \sim \Gamma(a_{\alpha_{z}},b_{\alpha_{z}}).  
\end{gathered}
\end{equation} 

\begin{figure}[H]
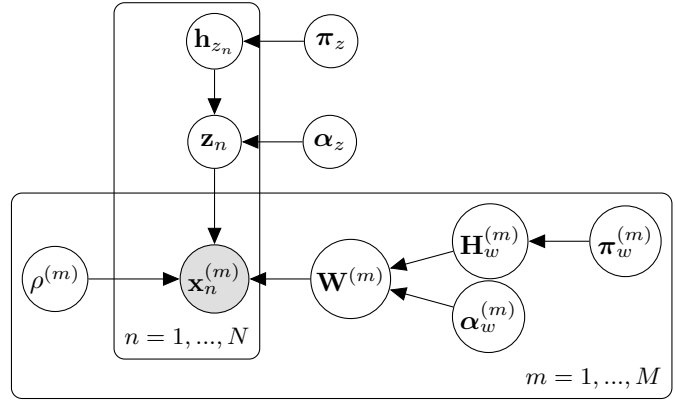

\centering
        \tikz{  
            \node[obs] (x) {$\mathbf{x}_{n}^{(m)}$};%
            \node[latent,above= of x] (z) {$\mathbf{z}_{n}$}; %
            \node[latent,right= 0.8 of z] (alz) {$\boldsymbol{\alpha}_{z}$};
            \node[latent,above= 0.6 of z] (hz) {$\mathbf{h}_{z_{n}}$};
            \node[latent,right= 0.8 of hz] (piz) {$\boldsymbol{\pi}_{z}$}; 
            \node[latent,left= 1.2 of x] (t) {$\rho^{(m)}$}; %
            \node[latent,right= 0.8 of x] (w) {$\mathbf{W}^{(m)}$}; %
            \node[latent,right= 0.8 of w, yshift=-0.5cm] (alw) {$\boldsymbol{\alpha}_{w}^{(m)}$};
            \node[latent,right= 0.8 of w, yshift=0.5cm] (hw) {$\mathbf{H}_{w}^{(m)}$};
            \node[latent,right= 0.8 of hw] (piw) {$\boldsymbol{\pi}_{w}^{(m)}$}; 
            \plate{plate1} {(x)(z)(hz)} {$n = 1,...,N$}; %
            \plate{plate2} {(t)(x)(w)(alw)(hw)(piw)} {$m = 1,...,M$}; %
            \edge {z} {x}
            \edge {w} {x}
            \edge {t} {x}
            \edge {alw} {w}
            \edge {hw} {w}
            \edge {piw} {hw}
            \edge {alz} {z}
            \edge {hz} {z}
            \edge {piz} {hz}
        }
\caption{Graphical representation of sparse GFA using spike-and-slab priors.} 
\label{fig:sGFA_spike_graph}            
\end{figure}

Assuming the priors described above, sparse GFA (displayed in Fig. \ref{fig:sGFA_spike_graph}) can identify subgroups in data that share common characteristics (e.g., brain-behaviour associations), which may be described as associations between subsets of features of two or more data modalities. The model parameters and latent variables are inferred using Gibbs sampling \citep{Bunte2016}.

\subsection{Sparse GFA using horseshoe priors}
\label{Methods_sGFA_hsprior}
The horseshoe prior \citep{Carvalho2009} is a popular shrinkage prior applied to Bayesian regression coefficients ($\boldsymbol{\beta} = (\beta_{1},...,\beta_{D})^T$, where $D$ is the number of features) that ensures that small coefficients are heavily shrunk towards zero, while large coefficients remain large. However, this property might be harmful in practice when the coefficients are weakly identified. \citet{Piironen2017} proposed a regularised extension of the horseshoe before ensuring that large $\beta$s are shrunk at least by a small amount. These priors are often termed global-local shrinkage priors, because there is a global parameter $\tau$ that shrinks all coefficients towards zero, while the local parameters $lambda$ allow some of these to escape complete shrinkage. The regularised horseshoe prior is defined as follows \citep{Piironen2017}:
\begin{equation} \label{eq7}
\begin{gathered}
	\beta_{j}|\lambda_{j},c,\tau \sim \mathcal{N}\left(0,\tfrac{c^{2} \tau^{2} \lambda_{j}^{2}}{c^{2} + \tau^{2}\lambda_{j}^{2}}\right), \quad \lambda_{j} \sim C^{+}(0,1), \quad j=1,...,D, \\
	c^{2} \sim \mathrm{InvGamma}(\nu/2, \nu s^{2}/2), \quad \tau \sim C^{+}(0,\tau_{0}^{2}), \quad \tau_{0} = \frac{p_{0}}{D - p_{0}} \frac{\sigma}{\sqrt{N}},  
\end{gathered}
\end{equation}
where $p_{0}$ is the prior guess of the number of relevant features, $\sigma$ is the noise standard deviation, $N$ is the number of samples and $C^{+}$ represents a half-Cauchy distribution (which is a Cauchy distribution truncated to only have non-zero probability density for values greater than or equal to the location of the peak). The inverse-Gamma distribution over $c^{2}$ corresponds to a Student-t$_{\nu}$($0,s^{2}$) (with $\nu$ degrees of freedom and scale $s^{2}$ - we set $\nu = 4$ and $s=2$, as in \citet{Piironen2017}) slab for coefficients far from zero. If the degrees of freedom $\nu$ are small enough, the prior will have heavy tails that ensure robust shrinkage of the large coefficients. In this way, the local parameters $\lambda$ cause small coefficients to shrink close to zero, while also regularising the largest ones. For more details on shrinkage priors for Bayesian regression, see \citet{Piironen2017} and \citet{VanErp2019}.

\begin{figure}[H]
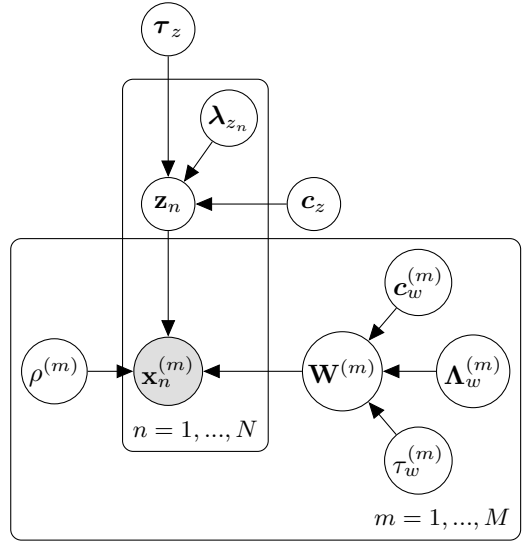

\centering
        \tikz{  
            \node[obs] (x) {$\mathbf{x}_{n}^{(m)}$};%
            \node[latent,above= 1.4 of x] (z) {$\mathbf{z}_{n}$}; %
            \node[latent,above= 1.6 of z] (tz) {$\boldsymbol{\tau}_{z}$};
            \node[latent,above= 0.4 of z, xshift= 0.8cm] (lz) {$\boldsymbol{\lambda}_{z_{n}}$};
            \node[latent,right= 1.2 of z] (cz) {$\boldsymbol{c}_{z}$};
            \node[latent,left= 0.6 of x] (s) {$\rho^{(m)}$}; %
            \node[latent,right= 1.3 of x] (w) {$\mathbf{W}^{(m)}$}; %
            \node[latent,below= 0.2 of w, xshift=1cm] (tw) {$\tau_{w}^{(m)}$};
            \node[latent,right= 0.7 of w] (lw) {$\mathbf{\Lambda}_{w}^{(m)}$};
            \node[latent,above= 0.2 of w, xshift=1cm] (cw) {$\boldsymbol{c}_{w}^{(m)}$}; 
            \plate{plate1} {(x)(z)(lz)} {$n = 1,...,N$}; %
            \plate{plate2} {(s)(x)(w)(tw)(lw)(cw)} {$m = 1,...,M$}; %
            \edge {z} {x}
            \edge {w} {x}
            \edge {s} {x}
            \edge {tw} {w}
            \edge {lw} {w}
            \edge {cw} {w}
            \edge {tz} {z}
            \edge {lz} {z}
            \edge {cz} {z}
        }
\caption{Graphical representation of sparse GFA using regularised horseshoe priors.}        
\label{fig:sGFA_hs_graph}            
\end{figure}

We replaced the spike-and-slab priors of sparse GFA proposed by \citet{Bunte2016} with regularised horseshoe priors (Fig. \ref{fig:sGFA_hs_graph}). The priors over the loading matrices ($\mathbf{W}^{(m)} \in \mathbb{R}^{D_{m} \times K}$, where $D_{m}$ is the number of features in the $m$th view and $K$ is the number of factors) are defined as follows:  
\begin{equation} \label{eq8}
\begin{gathered}
    w_{j,k}^{(m)}|\lambda_{w_{j,k}}^{(m)},c_{w_{k}}^{(m)},\tau_{w}^{(m)}
    \sim \mathcal{N}\left(0, \frac{(c_{w_{k}}^{(m)})^{2} (\tau_{w}^{(m)})^{2} (\lambda_{w_{j,k}}^{(m)})^{2}}{(c_{w_{k}}^{(m)})^{2} + (\tau_{w}^{(m)})^{2}(\lambda_{w_{j,k}}^{(m)})^{2}}\right), \quad j=1,...,D_{m}, \ k=1,...,K,\\
    \quad \lambda_{w_{j,k}}^{(m)} \sim C^{+}(0,1), \quad \tau_{w}^{(m)} \sim C^{+}(0,\tau_{0}^{2}), \quad \tau_{0}^{(m)} = \frac{p_{0}^{(m)}}{D_{m} - p_{0}^{(m)}} \frac{1}{\sqrt{N \rho^{(m)}}}, \quad \rho^{(m)} \sim \Gamma(a_{\rho},b_{\rho}),
\end{gathered}    
\end{equation}

where $p_{0}^{(m)}$ and $\rho^{(m)}$ are the prior guess of the number of relevant features and noise precision of the $m$th data modality, respectively. The prior over $(c_{w_{k}}^{(m)})^{2}$ is an inverse-Gamma distribution as defined in Equation \ref{eq7}. The regularised horseshoe prior allows different levels of sparsity across different data modalities, because a global shrinkage parameter $\tau_{w}^{(m)}$ is specified for each $m$th data modality. Moreover, as a $c$ value is sampled for each factor $k$ within a data modality, the prior implements sparsity over the loading matrices (i.e., $c_{w_{k}}^{(m)} \to 0$ leads to $\mathbf{w}_{k}^{(m)} \to 0$). A factor is deemed ``irrelevant'' if the loadings of that factor are pushed close to zero for all data modalities. Finally, the prior also implements feature-wise sparsity because some loadings escape shrinkage due to large local parameters in matrix $\boldsymbol{\Lambda}_{w}^{(m)}$. An analogous prior is used over the latent variables ($\mathbf{Z} \in \mathbb{R}^{K \times N}$) to include sample-wise sparsity: 
\begin{equation} \label{eq9}
\begin{gathered}
    z_{k,n}|\lambda_{z_{k,n}},c_{z_{k}},\tau_{z_{k}} \sim \mathcal{N}\left(0, \frac{c_{z_{k}}^{2} \tau_{z_{k}}^{2} \lambda_{z_{k,n}}^{2}}{c_{z_{k}}^{2} + \tau_{z_{k}}^{2}\lambda_{z_{k,n}}^{2}}\right), \quad n=1,...,N, \quad k=1,...,K,\\
    \tau_{z_{k}} \sim C^{+}(0,1), \quad \lambda_{z_{k,n}} \sim C^{+}(0,1),
\end{gathered}    
\end{equation}
where $c^{2}_{z_{k}}$ is drawn from the inverse-Gamma prior defined in Equation \ref{eq7}. Since a prior guess of the number of relevant samples is challenging to define, we set the scale of $\tau_{z_{k}}$ prior distribution to 1, as proposed in \citep{Carvalho2009}. The regularised horseshoe prior implements different sparsity levels across the latent factors by assuming different $\tau_{z_{k}}$. The interpretation of the effects of vector $\mathbf{c}_{z}$ and matrix $\boldsymbol{\Lambda}_{z}$ is equivalent to that described above for the loading matrices. Each data modality is then generated from the generative model in Equation \ref{eq1}.

The joint probability distribution of sparse GFA is then given by:
\begin{equation} \label{eq10}
\begin{aligned}
    & p(\mathbf{X},\mathbf{Z},\mathbf{W},\boldsymbol{\Lambda}_{w}, \boldsymbol{\tau}_{w}, \mathbf{c}_{w}, \boldsymbol{\Lambda}_{z}, \boldsymbol{\tau}_{z}, \mathbf{c}_{z}, \boldsymbol{\rho}) = \prod_{m=1}^{M} \bigg[ p(\mathbf{W}^{(m)}| \boldsymbol{\Lambda}^{(m)}_{w}, \boldsymbol{\tau}^{(m)}_{w}, \mathbf{c}^{(m)}_{w})  \\
    & p(\boldsymbol{\Lambda}^{(m)}_{w}) p(\boldsymbol{\tau}^{(m)}_{w}) p(\mathbf{c}^{(m)}_{w}) p(\rho^{(m)}) \prod_{n=1}^{N} \bigg( p(\mathbf{z}_{n}| \boldsymbol{\lambda}_{z_{n}}, \boldsymbol{\tau}_{z}, \mathbf{c}_{z}) p(\boldsymbol{\lambda}_{z_{n}}) \\
    & p(\mathbf{x}_{n}| \mathbf{z}_{n}, \mathbf{W}^{(m)}, \rho^{(m)}) \bigg) \bigg] p(\boldsymbol{\tau}_{z}) p(\mathbf{c}_{z}),
\end{aligned}    
\end{equation}
where $\boldsymbol{\lambda}_{z_{n}}$ is the $n$th column of $\boldsymbol{\Lambda}_{z}$.

\subsection{Model inference}
\label{Methods_MI}
The exact inference of sparse GFA is analytically intractable. Here, we used Hamiltonian Monte Carlo (HMC) \citep{Neal2011, Betancourt2013} to approximate the posterior distribution of sparse GFA. Briefly, sampling algorithms, such as Markov Chain Monte Carlo (MCMC) methods, approximate the posterior distribution by drawing samples from it to compute posterior estimates (e.g., posterior expectation and variance). MCMC methods use the properties of a Markov Chain to stochastically explore (using Markov transitions) the space close to the mode of the posterior distribution (i.e., the region with higher probability mass) \citep{Betancourt2018}. However, constructing appropriate transitions is very important to use these methods efficiently, which is often a challenging process. The Metropolis-Hastings algorithm \citep{Metropolis1953, Hastings1970} was proposed to mitigate this issue by considering a proposal and a correction step. In the former, a candidate sample is randomly generated from a proposal distribution (i.e., a probability distribution that will be compared to the posterior distribution), which is accepted if it falls into regions close to the high posterior probability mass, or rejected otherwise in the correction step \citep{Bishop2006, Betancourt2018}. However, this random ``walk'' is problematic in high dimensional spaces, where, due to its geometry, most of the proposal samples fall far away from the high probability regions. The rejection rate is therefore high, making the procedure very slow and with the possibility of several regions of the posterior distribution not being explored. It is possible to induce a larger acceptance rate by shrinking the covariance of the proposal distribution, but the transitions would be very small, which leads to extremely slow exploration. Moreover, even if the posterior distribution is well explored, the slow exploration yields large autocorrelations and imprecise estimates \citep{Betancourt2018}. Hamiltonian Monte Carlo (HMC) is an MCMC method that makes use of Hamiltonian dynamics to improve the exploration step and increase the acceptance rate in high dimensional spaces \citep{Betancourt2018}. In this way, approximate inference can be run more efficiently to compute good estimates of the model's parameters. 

In recent years, several probabilistic programming libraries, such as Stan \citep{Stan2019}, Edward \citep{Tran2016}, PyMC3 \citep{Salvatier2016} and NumPyro \citep{Phan2019}, have been developed to provide high-performance probabilistic modelling and inference. These libraries use, for instance, HMC and its extensions (e.g., the No-U-Turn Sampler (NUTS) \citep{Hoffman2014}) to run full Bayesian statistical inference efficiently even in high dimensional datasets. Here, we used NumPyro to implement sparse GFA. NumPyro is a Python library that uses automatic differentiation and end-to-end compilation to run HMC (with a NUTS implementation) on CPU/GPU, which allows efficient automatic inference, i.e., the user does not need to derive the inference equations. 

\subsection{Synthetic data}
\label{Synthetic_data}
To assess the ability of sparse GFA to identify underlying factors differently expressed across subgroups we first generate synthetic data with $K_{\mathrm{true}}=3$ factors: the first factor was mostly expressed by the first subgroup; the second factor was mostly expressed by the second subgroup; the third factor was similarly expressed across the three subgroups (as shown in Fig. \ref{fig:synthetic_results}). We generated a collection of datasets with $N=150$ samples (i.e., fifty samples per subgroup) and dimensions $D_{1}=60$, $D_{2} = 40$ and $D_{3}=20$ using the generative model in Equation \ref{eq1}, where the Gaussian priors over $\mathbf{W}^{(m)}$ and $\mathbf{z}_{n}$ are defined in Equations \ref{eq8} and \ref{eq9}, respectively, and the noise precisions were fixed: $\sigma^{(1)}=3$, $\sigma^{(2)}=6$ and $\sigma^{(3)}=4$. The parameters to construct the regularised horseshoe priors over $\mathbf{W}$ and $\mathbf{Z}$ were defined as follows: $(c_{w_{k}}^{(m)})^{2}$ and $c_{z_{k}}^{2}$ were sampled from the inverse-Gamma prior in Equation \ref{eq7} with $\nu=2$ and $s=2$ (as proposed by \citet{Piironen2017}); $\boldsymbol{\Lambda}_{w}^{(m)}$ and $\boldsymbol{\Lambda}_{z}$ were set manually, i.e., the values of the indices of the relevant features/samples were set to $100$ and the remainder to $0.01$ and $0.001$ respectively; $\tau_{w}^{(m)} = \tau_{0}^{(m)}$, where $p_{0}^{(m)} = D_{m}/3$ (i.e., one-third of the features in a given data modality were considered relevant); $\tau_{z_{k}}$ was set to 0.01. 

\subsection{GENFI data}
\label{GENFI_data}
We used cross-sectional brain structural MRI data and non-imaging data (e.g., psychometrics and other behavioural features) of 117 patients with genetic frontotemporal dementia from the Genetic Frontotemporal Dementia Initiative (GENFI) dataset (\url{https://www.genfi.org/}). The patients were recruited across 13 centres in the United Kingdom, Canada, Italy, the Netherlands, Sweden and Portugal. Sixty subjects were removed because they had more than one-third of the non-imaging features missing. The final 117 participants included 58 \textit{C9orf72}, 41 \textit{GRN} and 18 \textit{MAPT} mutation carriers. The structural MR images were parcellated into different cortical and subcortical regions using a multi-atlas segmentation propagation approach \citep{Cardoso2015} to calculate grey matter volumes of the left and right frontal, temporal, parietal, occipital, cingulate and insula cortices. An estimate of the volume of the cerebellum was also included. The subcortical volumes included left and right amygdala, caudate, hippocampus, pallidum, putamen and thalamus. In addition to regional volumetric features, volume asymmetry was calculated as proposed in \citet{Young2018}, i.e., the absolute value of the difference between the volumes of the right and left hemispheres, normalised by the total volume of both hemispheres. This asymmetry measure was log-transformed to improve normality. The total number of brain imaging features was 28 ($\mathbf{X}^{(1)} \in \mathbb{R}^{28 \times 117}$). For more details on the acquisition and preprocessing procedures of the structural MRI data, see \citet{Rohrer2015}. 

We included non-imaging features from informant questionnaires (which were completed by primary caregivers) assessing behaviour and disease severity, neuropsychological tasks (completed by patients) and medical assessments of disease severity. A brief description of each non-imaging feature is provided in Supplementary Table 1. We excluded features with more than $10\%$ of missing values. The remaining missing values were imputed by the median of the respective feature across subjects (as the percentage of missing data for most of the remaining features was below $1\%$). Four confounding variables were regressed from both data modalities: age, sex, education, and total intracranial volume. The total number of non-imaging features included was 34 ($\mathbf{X}^{(2)} \in \mathbb{R}^{34 \times 117}$). All features in both data modalities were standardised to have zero mean and unit variance. 

\subsubsection{Ethics statement}
Local ethics committees approved the study at each site, and all participants provided written informed consent. The study was conducted according to the Declaration of Helsinki.

\subsection{GFA and sparse GFA experiments}
To assess whether the proposed sparse GFA model has additional benefits, both GFA and sparse GFA were used to identify latent factors in the synthetic and GENFI data. In both approaches, we used the inferred latent variables to calculate the absolute latent scores, i.e., the absolute latent variable values in a given factor, and the factor contributions across the different subgroups, i.e., mean absolute latent scores within each subgroup, normalised to sum to one. 

The GFA and the sparse GFA were fitted using HMC with four sampling chains and 2,500 (synthetic data) and 6,000 (GENFI data) samples (the first 1,000 were discarded as warm-up) and randomly initialised five times. The best initialisation was selected to maximise the expected log joint posterior density. All sampling chains were initialised with $K=5$ and $K=20$ factors, for synthetic and GENFI data, respectively. The parameters of sparse GFA's prior distributions (Equation \ref{eq7} and Equation \ref{eq8}) were set as follows: $p_{0}^{(m)} = D_{m}/3$, $a_{\rho} = 1$, $b_{\rho} = 1$, $\nu = 4$ and $s=2$.

\subsubsection{Robustness of factors}
\label{Methods_robcomps}
To minimise the risk of obtaining factors that might have occurred by chance, we used a similar approach as in \citet{Bunte2016} to select factors consistent across the different sampling chains. As the factor indices can be arbitrarily permuted across different sampling chains, they must first be matched with factors similar to those of the other sampling chains. The factors are first averaged over the posterior samples within a chain (which can be done because the factor indices are stable within a chain) and then compared to the factors obtained in other sampling chains using cosine similarity. Two factors are considered similar if the cosine similarity is greater than 0.80. Finally, a factor is considered robust if obtained in more than half of the sampling chains.

\section{Results}
In this section, we present the results of the experiments with synthetic data (Section \ref{resSynt}) and GENFI data (Section \ref{resGENFI}) using sparse GFA and GFA. 

\subsection{Synthetic data}
\label{resSynt}
\begin{figure}[H]
\includegraphics[width=0.95\linewidth]{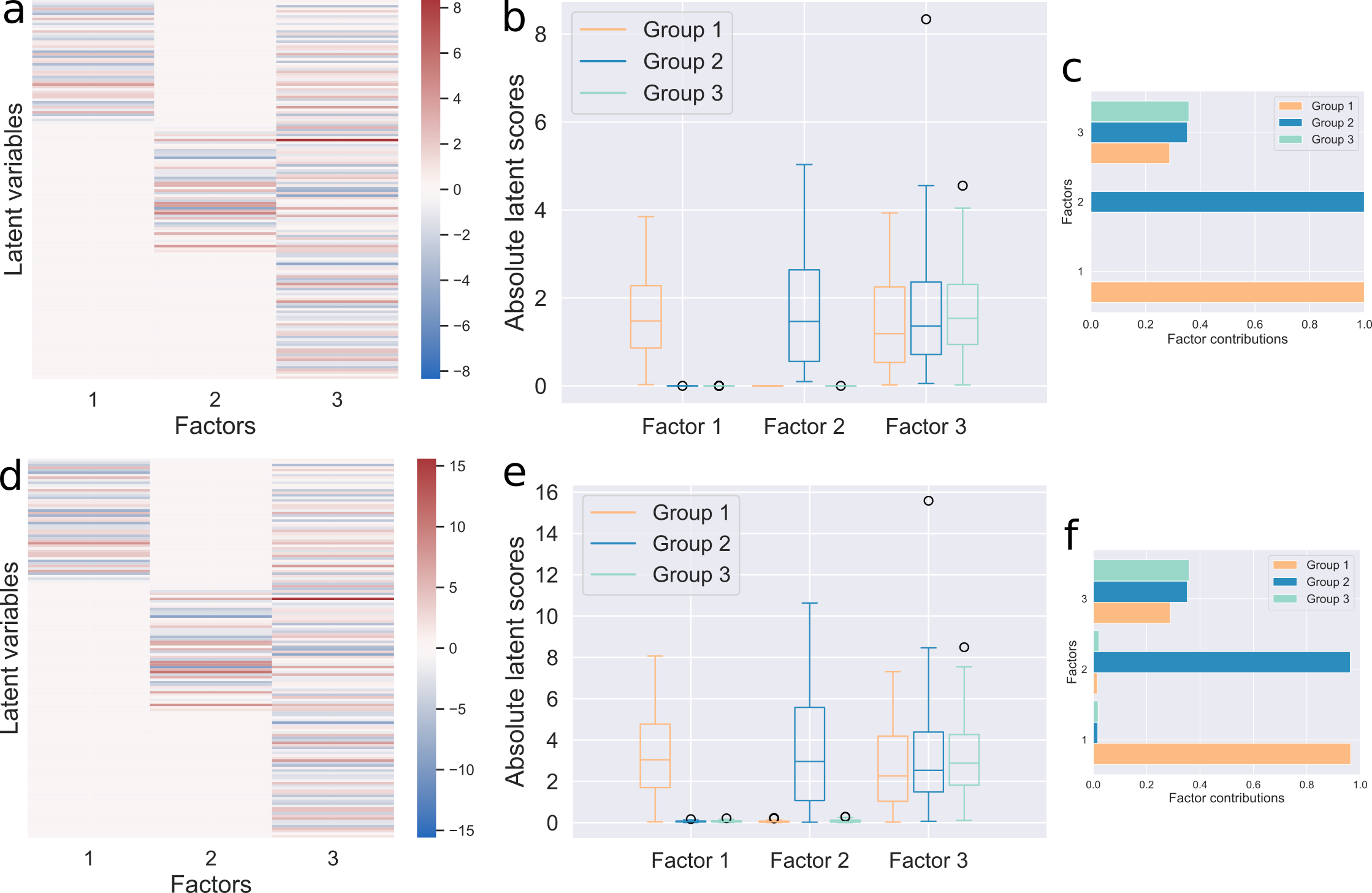}
\caption{\textbf{Synthetic data results using sparse GFA.} \textbf{(a)} True and \textbf{(d)} inferred latent variables by sparse GFA. Box plots showing the distribution of \textbf{(b)} true and \textbf{(e)} inferred absolute latent scores. \textbf{(c)} True and \textbf{(f)} inferred factor contributions across the different subgroups.}
\label{fig:synthetic_results} 
\end{figure}

\begin{figure}[H]
\includegraphics[width=0.95\linewidth]{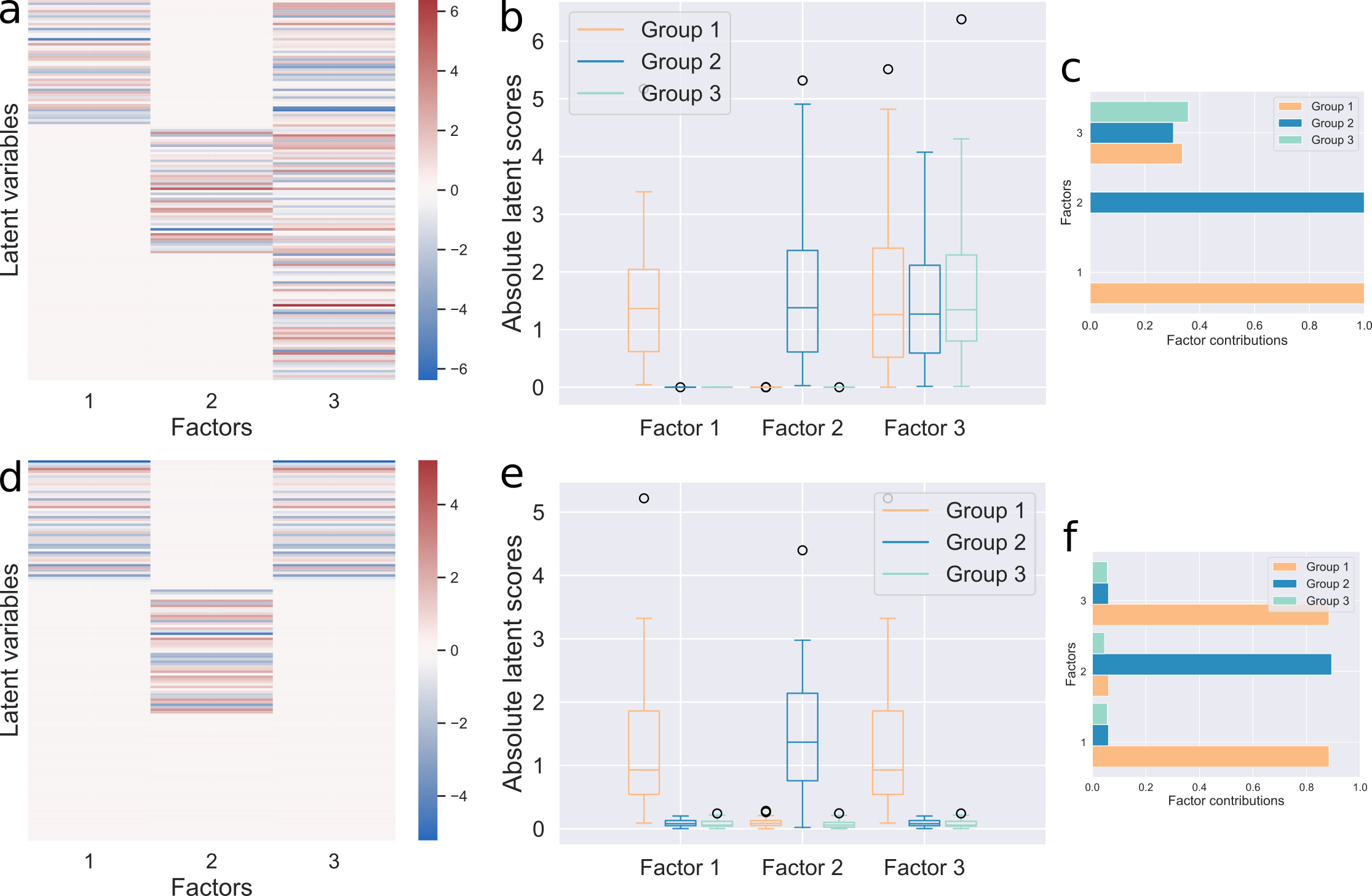}
\caption{\textbf{Synthetic data results using GFA.} \textbf{(a)} True and \textbf{(d)} inferred latent variables by GFA. Box plots showing the distribution of \textbf{(b)} true and \textbf{(f)} inferred absolute latent scores. \textbf{(c)} True and \textbf{(d)} inferred factor contributions of each factor. }
\label{fig:gfa_synthetic}
\end{figure}

Fig. \ref{fig:synthetic_results}a and Fig. \ref{fig:synthetic_results}b show the generated and inferred latent variables (represented by three distinct factors), respectively. The factors were correctly inferred by sparse GFA, i.e., the latent variables associated with each factor. Moreover, the number of factors was correctly estimated, i.e., the `irrelevant' factors were not considered robust using the approach described in Section \ref{Methods_robcomps}. Sparse GFA correctly estimated the distribution of the absolute latent scores and the factor contributions across the different subgroups (Fig. \ref{fig:synthetic_results}e-f). Both graphs show that the first/second factor was differently expressed in the first/second subgroup and the third factor was commonly expressed across the different subgroups. GFA identified the first and second factors but completely missed the factor shared across the subgroups (Fig. \ref{fig:gfa_synthetic}). When looking more carefully at the latent variables of the first and second factors (Fig. \ref{fig:gfa_synthetic}a-b), we can observe that GFA does not correctly infer the individual latent variables. 

\subsection{GENFI data}
\label{resGENFI}
\begin{figure}[H]
\includegraphics[width=\linewidth]{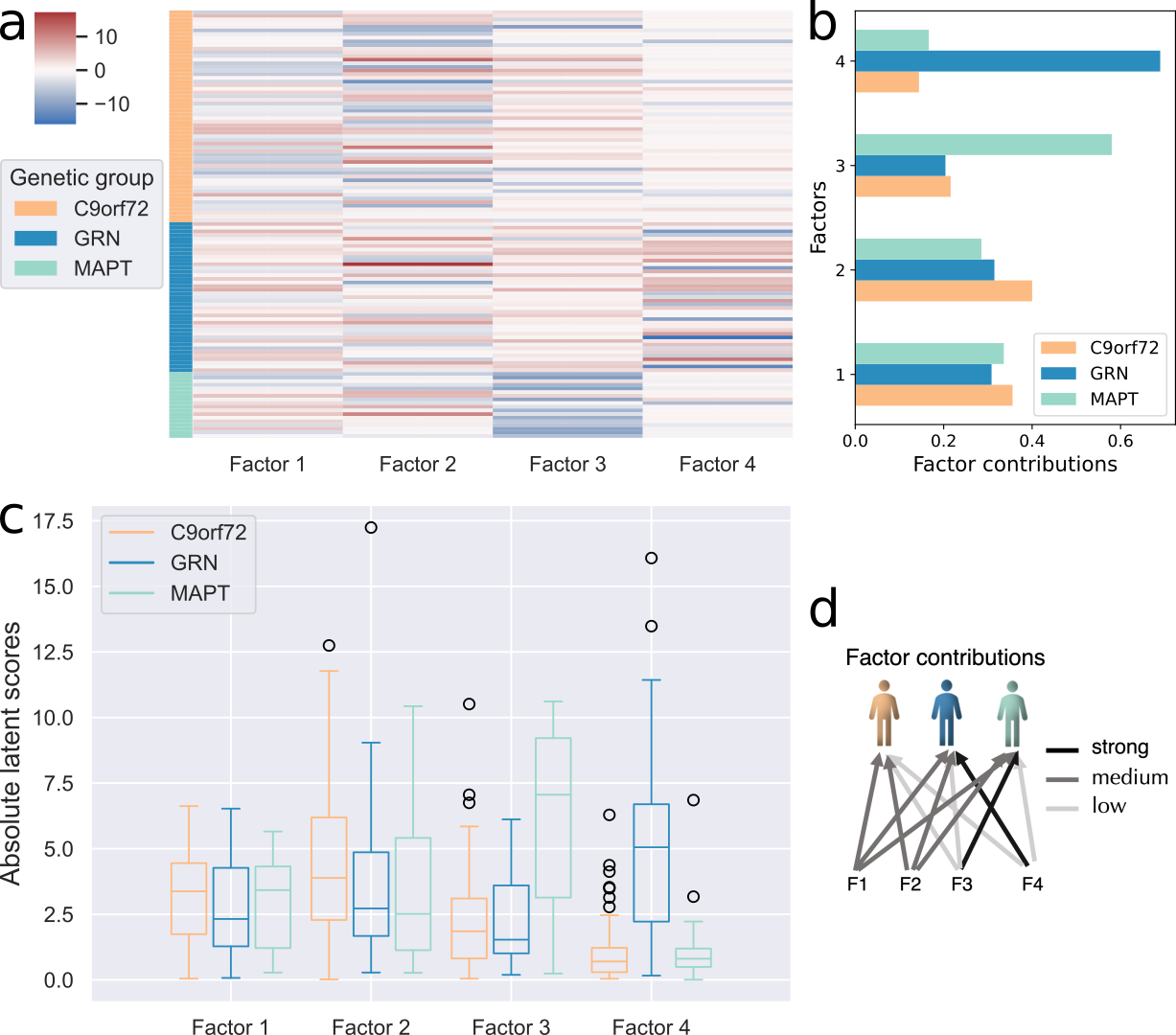}
\caption{\textbf{Robust factors identified by sparse GFA that explained most covariance in the data.} \textbf{(a)} Latent variables inferred by each factor grouped by genetic group. \textbf{(b)} and \textbf{(d)} Factor contributions for each genetic group. \textbf{(c)} Box plots showing the distribution of absolute latent scores of each factor.}
\label{fig:genfi_top4}
\end{figure}

Seventeen latent disease factors were considered robust using the approach described in Section \ref{Methods_robcomps}, which explained approximately $60\%$ of the covariance in the data (\nameref{sfig:all_factors}a). Here, we will carefully examine the four most relevant factors (which explained $\approx33\%$ of covariance), which were ranked by the amount of covariance they explained. The remaining factors are shown in the Supplementary Material (\nameref{sfig:all_factors}, \nameref{sfig:all_brain_loadings} and \nameref{sfig:all_NI_loadings}). Fig. \ref{fig:genfi_top4}b and \ref{fig:genfi_top4}c show the absolute latent scores and factor contributions within the three genetic groups of the four most relevant factors. The first and second factors were commonly expressed across the different groups (i.e., the mean of the absolute latent scores was similar across the different genetic groups), which was assessed by performing an F-test on the distributions of the absolute latent scores for each factor: $F=0.644 \ (p=0.527)$ for the first factor; $F=1.927 \ (p=0.150)$ for the second factor. At the same time, the third and fourth factors were differently expressed across the different groups ($F = 26.01 \ (p=4.943\times10^{-10})$ and $F = 38.311 \ (p=1.878\times10^{-13})$, respectively), showing a higher contribution for the \textit{MAPT} and \textit{GRN} groups, respectively (Fig. \ref{fig:genfi_top4}c). This difference can also be noted in the latent variables (Fig. \ref{fig:genfi_top4}a). In the third factor, the difference between the distribution of the absolute latent scores of the \textit{MAPT} group and those of the \textit{GRN} group ($T = 6.278$, \ $p=5.012\times10^{-8}$) and \textit{C9orf72} group ($T = 6.137$, \ $p=3.807\times10^{-8}$) were statistically significant (which was assessed by independent $t$-tests). In the fourth factor, the absolute latent scores of the \textit{GRN} group were statistically different when compared to those of the \textit{MAPT} ($T = 4.410$, \ $p=4.637\times10^{-5}$) and \textit{C9orf72} groups ($T = 8.001$, \ $p=2.677\times10^{-12}$). Figure \ref{fig:genfi_top4}d illustrates how the three different genetic groups can be characterized as a combination of the top four identified latent disease factors.

\begin{figure}[H]
\includegraphics[width=\linewidth]{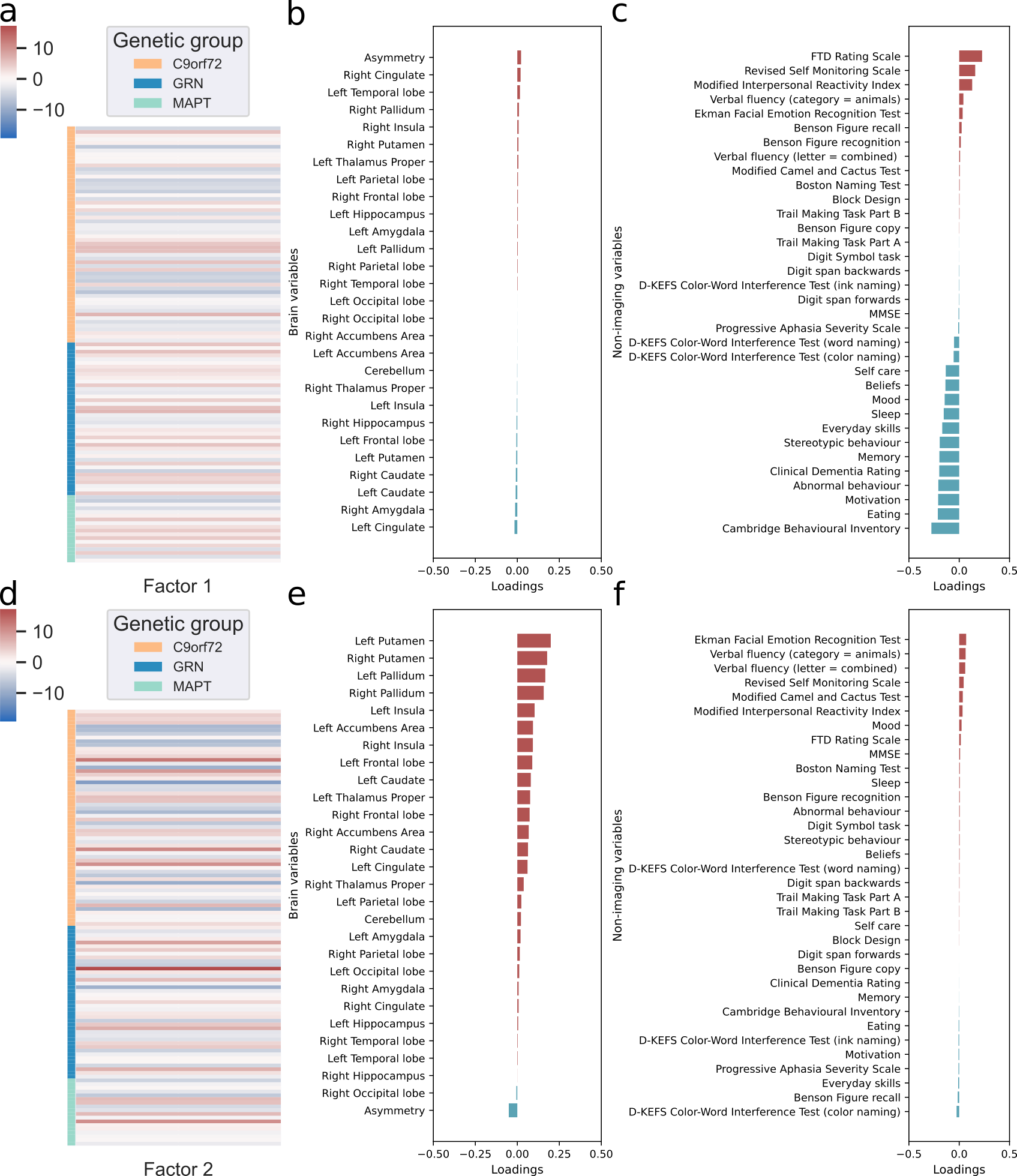}
\caption{\textbf{Factors with equal contribution to the three genetic groups.} Latent variables of the \textbf{(a)} first and \textbf{(d)} second factors grouped by genetic group. Brain loadings of the \textbf{(b)} first and \textbf{(e)} second factors. Non-imaging loadings of the \textbf{(c)} first and \textbf{(f)} second factors. More details about non-imaging features can be found in \nameref{stab:NI_vars}.}
\label{fig:non-specific_comps}
\end{figure}

The first latent disease factor (Fig. \ref{fig:non-specific_comps}a-c) captured associations across behavioural features and disease severity measurements, i.e., mostly within modality variance, with most brain features loading close to zero (Fig. \ref{fig:non-specific_comps}b). As it can be seen in Fig. \ref{fig:non-specific_comps}a, this factor is equally expressed across all genetic groups (confirmed by the factor contributions in Fig. \ref{fig:genfi_top4}c). Some individuals showed higher scores (shown in red in \nameref{sfig:data_comps}a) in multiple non-imaging features, e.g., the Cambridge Behavioural Inventory (which assesses general behaviour) and its 10 individual items (e.g., abnormal and stereotypic behaviour, memory and eating), and lower scores (shown in blue in \nameref{sfig:data_comps}a) related to disease severity (FTD rating scale), social behaviour (revised self-monitoring scale), and cognitive and emotional empathy (modified interpersonal reactivity index) - see \nameref{stab:NI_vars} for more details about non-imaging features. This factor highlights common behavioural and cognitive changes of individuals with frontotemporal dementia, where \textit{C9orf72} mutation carriers seem to be more affected than the \textit{GRN} and \textit{MAPT} and mutation carriers.

The second latent disease factor (Fig. \ref{fig:non-specific_comps}d-f) captured an association between symmetric subcortical grey matter changes (e.g., putamen, pallidum, caudate, thalamus and amygdala) and insula, with little contribution from the non-imaging features. This factor was also equally expressed across the entire population (Fig. \ref{fig:genfi_top4}c) and highlights grey matter volume changes across patients.

\begin{figure}[H]
\includegraphics[width=\linewidth]{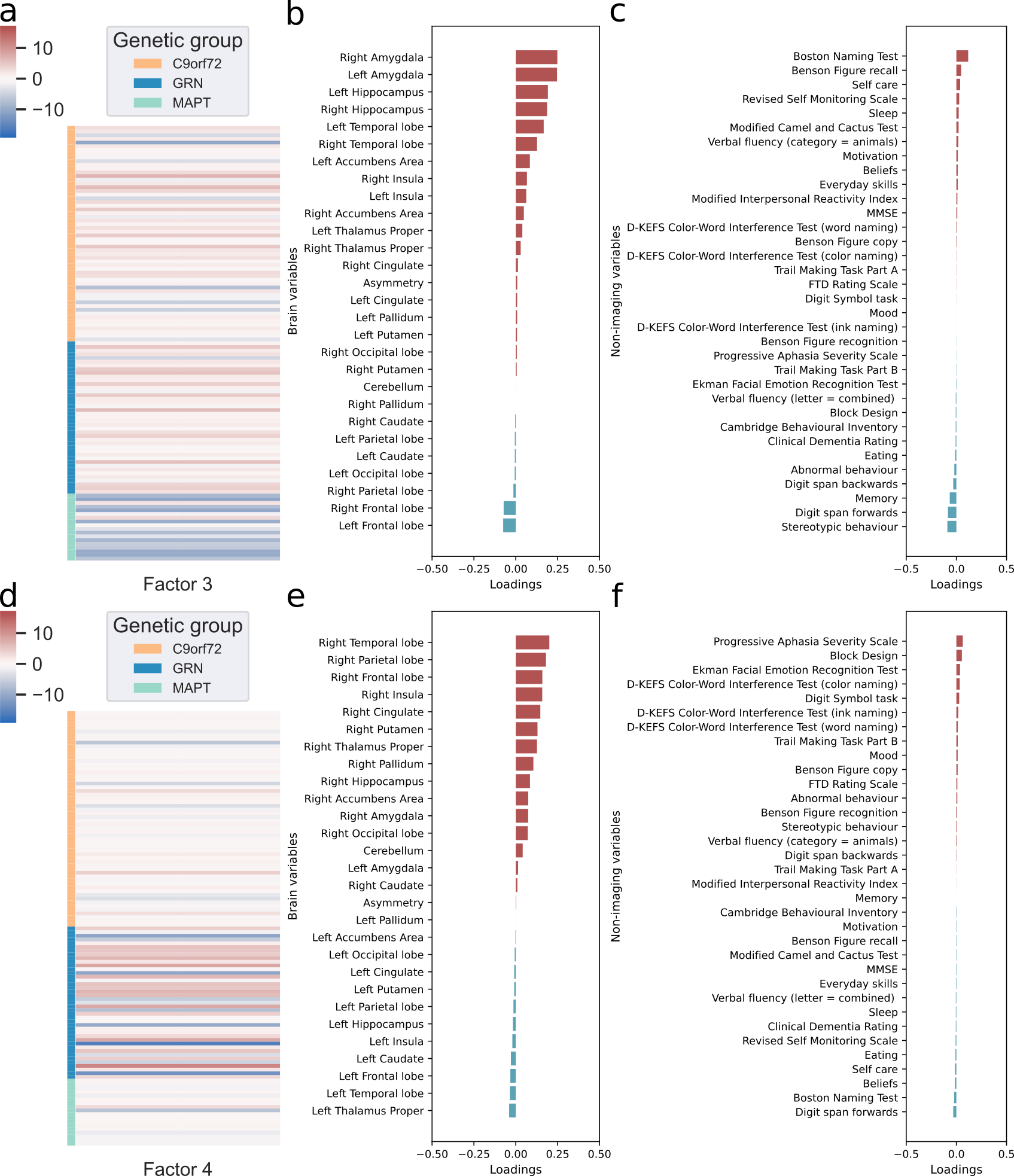}
\caption{\textbf{Factors differently expressed in the \textit{MAPT} and \textit{GRN} genetic groups} (third and fourth factors, respectively). Latent variables of the \textbf{(a)} third and \textbf{(d)} fourth factors grouped by genetic group. Brain loadings of the \textbf{(b)} third and \textbf{(e)} fourth factors. Non-imaging loadings of the \textbf{(c)} third and \textbf{(f)} fourth factors. More details about non-imaging features can be found in \nameref{stab:NI_vars}.}
\label{fig:specific_comps}
\end{figure}

The third latent disease factor (Fig. \ref{fig:specific_comps}a-c) captured a relationship between symmetric grey matter changes, in the right and left temporal lobes, amygdala, insula and hippocampus, and memory, stereotypic behaviour, the digit span forwards and Boston naming tasks. The \textit{MAPT} mutation carriers showed smaller grey matter volume in the regions mentioned above and the Boston naming task when compared to the \textit{C9orf72} and \textit{GRN} mutation carriers (shown in blue in \nameref{sfig:data_comps}c), and higher scores in memory, stereotypic behaviour and in the digit span forward task. As confirmed by the factor contributions, this factor was differently expressed in the \textit{MAPT} subgroup (Figure \ref{fig:genfi_top4}c), highlighting changes known to be associated with \textit{MAPT} mutation carriers.

The fourth latent disease factor (Fig. \ref{fig:specific_comps}d-f) showed an association between asymmetric grey matter changes (i.e., positive loadings for the right brain regions and negative loadings for the left brain regions) and speech, language (progressive aphasia severity scale), visuospatial skills (block design task), facial emotion recognition and executive function (Stroop task). This factor showed a greater contribution to the \textit{GRN} genetic group, indicating that it was differently expressed in this group (Figure \ref{fig:genfi_top4}c) and that it identified typical grey matter volume, behavioural and cognitive changes of these individuals.

\begin{figure}[H]
\includegraphics[width=\linewidth]{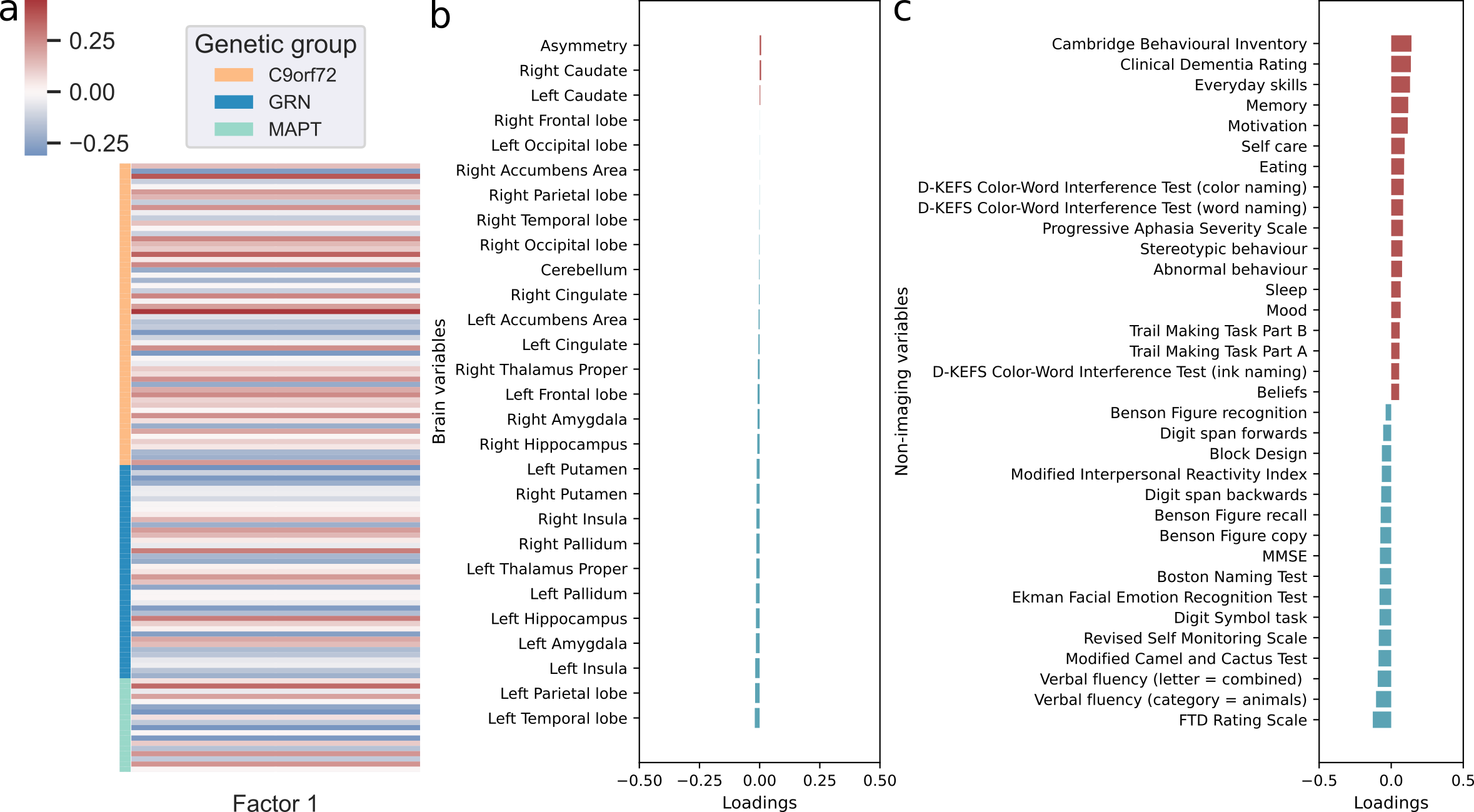}
\caption{\textbf{Robust factor identified by GFA in the GENFI data}. \textbf{(a)} Latent variables grouped by genetic group. \textbf{(b)} Brain and \textbf{(c)} non-imaging loadings. More details about non-imaging features can be found in \nameref{stab:NI_vars}.}
\label{fig:gfa_genfi}
\end{figure}

GFA was not able to identify any latent disease factor that was differently expressed across the genetic groups (see \ref{fig:gfa_genfi}). Indeed only one latent disease factor was found robust, which mostly captured overall behavioural and cognitive changes across the entire sample. In summary,  the results on synthetic and GENFI data show that the sparse extension of GFA allows us to identify latent disease factors differently expressed in subgroups, whereas this is not possible with vanilla GFA.

\section{Discussion}
In this study, we proposed a sparse extension of GFA using regularised horseshoe priors to simultaneously impose sparsity over the data modalities and features, as well as sparsity over the samples, to improve model interpretability and identify latent factors differently expressed across the sample. We show how the proposed sparse GFA approach can be applied to identify subgroup-specific and subgroup-common multimodal latent disease factors in patient groups.

We showed in synthetic data that our implementation of sparse GFA can correctly infer the latent structure of each subgroup, estimate the distribution of the latent variables, as well as calculate the factor contributions (Fig. \ref{fig:synthetic_results}), whereas GFA failed to do so in particular for the third factor (Fig. \ref{fig:gfa_synthetic}). In the GENFI data, our sparse GFA implementation identified several latent disease factors capturing associations between brain structure and non-imaging features (i.e., behavioural features, disease severity and cognitive assessments). These provided important information about the disease profile of the different genetic groups and were in line with previous findings. 

The factors that explained most covariance in the data were similarly expressed across all genetic groups (Fig. \ref{fig:non-specific_comps}) with some differences across the groups. The first latent disease factor showed a contrasting pattern between the \textit{C9orf72} mutation carriers (which is known to be a heterogeneous group \citep{Mahoney2012}) and the individuals in the other genetic groups (Fig. \ref{fig:non-specific_comps}a-c). The \textit{C9orf72} mutation carriers showed worse overall behaviour, and worse social cognition than the other genetic groups (\nameref{sfig:data_comps}a). These results are in keeping with previous studies where earlier changes in empathy and social emotional cognition are seen in C9orf72 mutation carriers \citep{Russell2020, Franklin2021} compared with the other genetic groups. 

The second most relevant latent disease factor (Fig. \ref{fig:non-specific_comps}d-f) shows characteristics of genetic FTD in terms of executive function, attention processing and social cognition, and grey matter volume changes in the left and right subcortical structures as well as the insula. Early subcortical involvement, particularly in \textit{C9orf72} mutation carriers, aligns with previous work \citep{Bocchetta2021}. Bocchetta and colleagues found that in all three groups, subcortical involvement can be identified early in the disease course, particularly in \textit{C9orf72} and \textit{MAPT} mutation carriers, involving volume changes in thalamic subnuclei, cerebellum, hippocampus, amygdala, and hypothalamus. \textit{C9orf72} mutation carriers were found to show the earliest and most widespread changes, including the thalamus, basal ganglia and medial temporal lobe \citep{Rohrer2015}. 

Interestingly, the third and fourth latent disease factors were most strongly expressed in the more homogeneous patient subgroups \citep{Mahoney2012}, namely the \textit{MAPT} (third latent disease factor, Fig. \ref{fig:specific_comps}a-c) and the \textit{GRN}  (fourth latent disease factor, Fig. \ref{fig:specific_comps}d-f) carriers. The third latent disease factor showed that symptomatic \textit{MAPT} carriers had a symmetrical pattern of atrophy involving the temporal lobe volume (\nameref{sfig:data_comps}c), as well as in the amygdala and hippocampus, as reported in previous findings \citep{Rohrer2010, Rohrer2013, Whitwell2009}. The \textit{MAPT} mutation carriers had worse object naming than other genetic groups, which has been found in previous work \citep{Rohrer2015}, and can even dissociate this genetic group from the others at a presymptomatic stage \citep{Bouzigues2022}. 

The fourth latent disease factor showed that symptomatic \textit{GRN} mutation carriers had a prominent asymmetrical pattern of atrophy in either the left or the right frontal, temporal and parietal lobes, which replicates previous findings \citep{Rohrer2010, Mahoney2012, Gordon2016}. These individuals had worse executive function, attention, language and - not as impaired - on certain everyday behaviours (\nameref{sfig:data_comps}d). These results align with previous literature that suggests that symptomatic \textit{GRN} mutation carriers show a diverse range of behavioural, cognitive and language deficits \citep{Rohrer2010}. Interestingly, within the \textit{GRN} group, some mutation carriers showed an opposite pattern on the same brain and non-imaging features, i.e., the positive and negative values were flipped (see \nameref{sfig:data_comps}d). Language problems are more common (in left-sided cases) in this mutation group than in others. 

These findings demonstrate that our implementation of sparse GFA can successfully identify data-driven multimodal latent disease factors that reveal important disease profiles common across all patients and disease profiles specific to subgroups of patients. Furthermore, sparse GFA can help understand disease heterogeneity at the single subject level, i.e. the latent disease factors can also provide a disease profile at the subject level. One can easily visualise the distribution of the different brain and non-imaging scores over the subjects by mapping the latent disease factor to the data space, i.e. $\mathbf{X}_{k} = \mathbf{w}_{(:,k)} \mathbf{z}_{(k,:)}$, $k = 1,...,K$ (\nameref{sfig:data_comps}). For instance, in \nameref{sfig:data_comps}c it is possible to see that most \textit{MAPT} mutation carriers present values below the sample mean (i.e., in blue) for the right and left amygdala, hippocampus and temporal lobe, indicating that these individuals have lower grey matter volume in these regions than the remaining participants, which aligns well with previous studies \citep{Rohrer2010, Rohrer2013, Whitwell2009}.

Sparse GFA is particularly suited to addressing a long-standing issue in mental health research, that effective treatment selection for mental health disorders is confounded by heterogeneity within traditional diagnostic categories and comorbidity between them \citep{Kessler2005}. By incorporating sparsity over samples, sparse GFA is equipped to identify latent dimensions selectively expressed within a heterogenous patient population, that represent potential (differential) mechanisms of disease. Applying sparse GFA to a dataset comprising multiple mental health disorders (in a ‘transdiagnostic’ approach) has the potential to reveal latent disease dimensions that cut across diagnostic categories, highlighting common potential disease mechanisms that may also explain the extensive comorbidity between disorders. In this way, sparse GFA can help refine psychiatric nosology in a data-driven and biologically-informed approach, paving the way towards ‘precision medicine’ in mental healthcare \citep{Bzdok2017}.

While there has been recent interest in finding ‘subtypes’ or subgroups within diagnostic categories such as depression \citep{Drysdale2017} and schizophrenia \citep{Clementz2016}, we note that such subgroups may be sufficiently fuzzy and overlapping that a dimensional approach (rather than categorical) may better describe variation in these patients. It is an open empirical question as to whether continuous dimensions or categorical subgroups offer more clinical utility in guiding treatment selection or predicting prognostic outcomes \citep{Mihalik2020b}. Indeed, categorical subgroups in depression have proven difficult to replicate \citep{Dinga2019}, perhaps due to the inherent instability of clustering methods.

A sparse GFA solution represents each patient as expressing a mixture of different multimodal latent dimensions, which combine to produce the individual’s set of biomarkers. In this way, sparse GFA has a natural alignment to the influential RDoC framework for studying mental health disorders, which emphasises a continuous dimensional approach in understanding heterogeneity and comorbidity \citep{Insel2010, Morris2022}. RDoC encourages fusing multiple data modalities (`units of analysis'), which sparse GFA readily accommodates. Indeed, the potential for identifying novel latent disease dimensions using sparse GFA depends on incorporating rich multimodal data, allowing for connections between different data modalities that might have been previously overlooked by clinician and researcher expertise.

In this study, we have shown an application of our sparse GFA implementation on solely two data modalities. However, its applicability extends seamlessly to encompass various data modalities, such as multiple imaging modalities (e.g., structural and functional MRI) or alternative data types (e.g., omics data). The fusion of additional data modalities holds promise for identifying additional latent disease factors of greater significance, potentially elucidating disease comorbidities and underlying heterogeneity within patients. Future studies could apply sparse GFA to large multimodal population datasets, such as the ABCD study \citep{Volkow2017} or UK Biobank \citep{Miller2016}, to uncover latent factors in the general population potentially related to psychopathology or other diseases. Given that the proposed sparse GFA implementation uses sampling algorithms for inference, it might not scale well for high-dimensional datasets. However, as sparse GFA was implemented within a probabilistic programming framework, inference can be scaled up using variational inference algorithms (which are available in these probabilistic programming libraries).

In summary, sparse GFA is an unsupervised machine method that can uncover interpretable latent disease factors to provide insights into the dimensions of disease and improve the characterisation of disease subgroups. The sparse GFA approach enables us to move towards identifying biologically defined patient subgroups that can be described by several disease processes in grated degrees as conceptualized in \citet{Bzdok2017}. Finally, our sparse GFA implementation leverages the powerful programming language algorithms to be easily extended to more complex models and applied to other neuroimaging tasks or fields of research, such as mental health.

\section*{Data and code availability}
GFA and sparse GFA were implemented in Python 3.9.1 and can be found at \url{https://github.com/ferreirafabio80/sgfa} (to be made public upon paper acceptance). All study data, including raw and analysed data, and materials will be available upon reasonable request.

\section*{Acknowledgements}
FSF was supported by Fundação para a Ciência e a Tecnologia (PhD fellowship No. SFRH/BD/120640/2016). JM-M was supported by the Wellcome Trust under Grant No. WT102845/Z/13/Z. The Wellcome Centre for Human Neuroimaging is supported by core funding from the Wellcome Trust (203147/Z/16/Z).
  
\section*{Declaration of Competing Interests}
The authors do not have any conflicts of interest to disclose.

\section*{Author Contributions}
Fabio S. Ferreira: conceptualization, methodology, formal analysis, investigation, data curation, visualization, writing – original draft. John Ashburner: conceptualization, writing – review \& editing, supervision. Arabella Bouzigues: writing – review \& editing. Chatrin Suksasilp: writing – review \& editing. Jonathan D. Rohrer: writing – review \& editing, data curation and collection. Samuel Kaski: conceptualization, writing – review \& editing, supervision. Janaina Mourao-Miranda: conceptualization, writing – review \& editing, supervision, project administration, funding acquisition. Members of the GENFI consortium recruited patients and collected and pre-processed data. All authors contributed to reviewing and editing the manuscript.

\newpage

\section*{Supplementary Material}
\paragraph*{Appendix 1}
\label{appendix_GENFI}
List of collaborators in the GENFI consortium

\begin{longtable}[H]{p{0.23\textwidth}|p{0.7\textwidth}}
\toprule
\textbf{\small Author} & \textbf{ \small Affiliation} \\
\midrule
\small Rhian Convery &  \small Department of Neurodegenerative Disease, Dementia Research Centre, UCL Queen Square Institute of Neurology, London, UK \\
\hline
\small Martina Bocchetta &  \small Department of Neurodegenerative Disease, Dementia Research Centre, UCL Queen Square Institute of Neurology, London, UK \\
\hline
\small David Cash &  \small Department of Neurodegenerative Disease, Dementia Research Centre, UCL Queen Square Institute of Neurology, London, UK \\
\hline
\small Sophie Goldsmith &  \small Department of Neurodegenerative Disease, Dementia Research Centre, UCL Queen Square Institute of Neurology, London, UK \\
\hline
\small Kiran Samra &  \small Department of Neurodegenerative Disease, Dementia Research Centre, UCL Queen Square Institute of Neurology, London, UK \\
\hline
\small David L. Thomas &  \small Neuroimaging Analysis Centre, Department of Brain Repair and Rehabilitation, UCL Institute of Neurology, Queen Square, London, UK \\
\hline
\small Thomas Cope &  \small Cambridge University Hospitals NHS Trust, Cambridge UK \\
\hline
\small Maura Malpetti &  \small Department of Clinical Neurosciences, University of Cambridge, Cambridge, UK \\
\hline
\small Antonella Alberici &  \small Centre for Neurodegenerative Disorders, Department of Clinical and Experimental Sciences, University of Brescia, Brescia, Italy \\
\hline
\small Enrico Premi &  \small Stroke Unit, ASST Brescia Hospital, Brescia, Italy \\
\hline
\small Roberto Gasparotti &  \small Neuroradiology Unit, University of Brescia, Brescia, Italy \\
\hline
\small Emanuele Buratti &  \small ICGEB Trieste, Italy \\
\hline
\small Valentina Cantoni &  \small Centre for Neurodegenerative Disorders, Department of Clinical and Experimental Sciences, University of Brescia, Brescia, Italy \\
\hline
\small Andrea Arighi &  \small Fondazione IRCCS Ca’ Granda Ospedale Maggiore Policlinico, Neurodegenerative Diseases Unit, Milan, Italy\\
\hline
\small Chiara Fenoglio &  \small University of Milan, Centro Dino Ferrari, Milan, Italy\\
\hline
\small Vittoria Borracci &  \small Fondazione IRCCS Ca’ Granda Ospedale Maggiore Policlinico, Neurodegenerative Diseases Unit, Milan, Italy\\
\hline
\small Maria Serpente &  \small Fondazione IRCCS Ca’ Granda Ospedale Maggiore Policlinico, Neurodegenerative Diseases Unit, Milan, Italy\\
\hline
\small Tiziana Carandini &  \small Fondazione IRCCS Ca’ Granda Ospedale Maggiore Policlinico, Neurodegenerative Diseases Unit, Milan, Italy\\
\hline
\small Emanuela Rotondo &  \small Fondazione IRCCS Ca’ Granda Ospedale Maggiore Policlinico, Neurodegenerative Diseases Unit, Milan, Italy\\
\hline
\small Giacomina Rossi &  \small Fondazione IRCCS Istituto Neurologico Carlo Besta, Milano, Italy\\
\hline
\small Giorgio Giaccone &  \small Fondazione IRCCS Istituto Neurologico Carlo Besta, Milano, Italy\\
\hline
\small Giuseppe Di Fede &  \small Fondazione IRCCS Istituto Neurologico Carlo Besta, Milano, Italy\\
\hline
\small Paola Caroppo &  \small Fondazione IRCCS Istituto Neurologico Carlo Besta, Milano, Italy\\
\hline
\small Sara Prioni &  \small Fondazione IRCCS Istituto Neurologico Carlo Besta, Milano, Italy\\
\hline
\small Veronica Redaelli &  \small Fondazione IRCCS Istituto Neurologico Carlo Besta, Milano, Italy\\
\hline
\small David Tang-Wai &  \small The University Health Network, Krembil Research Institute, Toronto, Canada\\
\hline
\small Ekaterina Rogaeva &  \small Tanz Centre for Research in Neurodegenerative Diseases, University of Toronto, Toronto, Canada\\
\hline
\small Johanna Krüger &  \small Research Unit of Clinical Medicine, Neurology, University of Oulu, Oulu, Finland\\
\hline
\small Miguel Castelo-Branco &  \small Faculty of Medicine, ICNAS, CIBIT, University of Coimbra, Coimbra, Portugal.\\
\hline
\small Morris Freedman &  \small Baycrest Health Sciences, Rotman Research Institute, University of Toronto, Toronto, Canada\\
\hline
\small Ron Keren &  \small The University Health Network, Toronto Rehabilitation Institute, Toronto, Canada\\
\hline
\small Sandra Black &  \small Sunnybrook Health Sciences Centre, Sunnybrook Research Institute, University of Toronto, Toronto, Canada\\
\hline
\small Sara Mitchell &  \small Sunnybrook Health Sciences Centre, Sunnybrook Research Institute, University of Toronto, Toronto, Canada\\
\hline
\small Christen Shoesmith &  \small Department of Clinical Neurological Sciences, University of Western Ontario, London, Ontario, Canada\\
\hline
\small Robart Bartha &  \small Department of Medical Biophysics, The University of Western Ontario, London, Ontario, Canada\\
\hline
\small Rosa Rademakers &  \small Center for Molecular Neurology, University of Antwerp\\
\hline
\small Jackie Poos &  \small Department of Neurology, Erasmus Medical Center, Rotterdam, Netherlands\\
\hline
\small Janne M. Papma &  \small Department of Neurology, Erasmus Medical Center, Rotterdam, Netherlands\\
\hline
\small Lucia Giannini &  \small Department of Neurology, Erasmus Medical Center, Rotterdam, Netherlands\\
\hline
\small Liset de Boer &  \small Department of Neurology, Erasmus Medical Center, Rotterdam, Netherlands\\
\hline
\small Julie de Houwer &  \small Department of Neurology, Erasmus Medical Center, Rotterdam, Netherlands\\
\hline
\small Rick van Minkelen &  \small Department of Clinical Genetics, Erasmus Medical Center, Rotterdam, Netherlands\\
\hline
\small Yolande Pijnenburg &  \small Amsterdam University Medical Centre, Amsterdam VUMC, Amsterdam, Netherlands\\
\hline
\small Benedetta Nacmias &  \small Department of Neuroscience, Psychology, Drug Research and Child Health, University of Florence, Florence, Italy\\
\hline
\small Camilla Ferrari &  \small Department of Neuroscience, Psychology, Drug Research and Child Health, University of Florence, Florence, Italy\\
\hline
\small Cristina Polito &  \small Department of Biomedical, Experimental and Clinical Sciences “Mario Serio”, Nuclear Medicine Unit, University of Florence, Florence, Italy\\
\hline
\small Gemma Lombardi &  \small Department of Neuroscience, Psychology, Drug Research and Child Health, University of Florence, Florence, Italy\\
\hline
\small Valentina Bessi &  \small Department of Neuroscience, Psychology, Drug Research and Child Health, University of Florence, Florence, Italy\\
\hline
\small Enrico Fainardi &  \small Neuroradiology Unit, Department of Experimental and Clinical Biomedical Sciences, University of Florence, Florence, Italy\\
\hline
\small Stefano Chiti &  \small Neuroradiology Unit, Department of Experimental and Clinical Biomedical Sciences, University of Florence, Florence, Italy\\
\hline
\small Mattias Nilsson &  \small Department of Clinical Neuroscience, Karolinska Institutet, Stockholm, Sweden\\
\hline
\small Henrik Viklund &  \small Karolinska University Hospital Huddinge\\
\hline
\small Melissa Taheri Rydell &  \small Department of Neurobiology, Care Sciences and Society; Center for Alzheimer Research, Division of Neurogeriatrics, , Bioclinicum, Karolinska Institutet, Solna, Sweden\\
\hline
\small Vesna Jelic &  \small Department of Neurobiology, Care Sciences and Society; Division of Clinical Geriatrics, Karolinska Institutet, Stockholm, Sweden\\
\hline
\small Linn Öijerstedt &  \small Department of Neurobiology, Care Sciences and Society; Center for Alzheimer Research, Division of Neurogeriatrics, , Bioclinicum, Karolinska Institutet, Solna, Sweden\\
\hline
\small Tobias Langheinrich &  \small Division of Neuroscience and Experimental Psychology, Wolfson Molecular Imaging Centre, University of Manchester, Manchester, UK\\
\hline
\small Albert Lladó &  \small Alzheimer’s disease and Other Cognitive Disorders Unit, Neurology Service, Hospital Clínic, Barcelona, Spain\\
\hline
\small Anna Antonell &  \small Alzheimer’s disease and Other Cognitive Disorders Unit, Neurology Service, Hospital Clínic, Barcelona, Spain\\
\hline
\small Jaume Olives &  \small Alzheimer’s disease and Other Cognitive Disorders Unit, Neurology Service, Hospital Clínic, Barcelona, Spain\\
\hline
\small Mircea Balasa &  \small Alzheimer’s disease and Other Cognitive Disorders Unit, Neurology Service, Hospital Clínic, Barcelona, Spain\\
\hline
\small Nuria Bargalló &  \small Imaging Diagnostic Center, Hospital Clínic, Barcelona, Spain\\
\hline
\small Sergi Borrego-Ecija &  \small Alzheimer’s disease and Other Cognitive Disorders Unit, Neurology Service, Hospital Clínic, Barcelona, Spain\\
\hline
\small Ana Verdelho &  \small Department of Neurosciences and Mental Health, Centro Hospitalar Lisboa Norte - Hospital de Santa Maria \& Faculty of Medicine, University of Lisbon, Lisbon, Portugal\\
\hline
\small Carolina Maruta &  \small Laboratory of Language Research, Centro de Estudos Egas Moniz, Faculty of Medicine, University of Lisbon, Lisbon, Portugal\\
\hline
\small Tiago Costa-Coelho &  \small Faculty of Medicine, University of Lisbon, Lisbon, Portugal\\
\hline
\small Gabriel Miltenberger &  \small Faculty of Medicine, University of Lisbon, Lisbon, Portugal\\
\hline
\small Alazne Gabilondo &  \small Cognitive Disorders Unit, Department of Neurology, Donostia University Hospital, San Sebastian, Gipuzkoa, Spain\\
\hline
\small Ioana Croitoru &  \small Instituto de Investigación Sanitaria Biogipuzkoa, Neurosciences Area, Group of Neurodegenerative Diseases, San Sebastian, Spain\\
\hline
\small Mikel Tainta &  \small Instituto de Investigación Sanitaria Biogipuzkoa, Neurosciences Area, Group of Neurodegenerative Diseases, San Sebastian, Spain\\
\hline
\small Myriam Barandiaran &  \small Cognitive Disorders Unit, Department of Neurology, Donostia University Hospital, San Sebastian, Gipuzkoa, Spain\\
\hline
\small Patricia Alves &  \small Instituto de Investigación Sanitaria Biogipuzkoa, Neurosciences Area, Group of Neurodegenerative Diseases, San Sebastian, Spain\\
\hline
\small Benjamin Bender &  \small Department of Diagnostic and Interventional Neuroradiology, University of Tübingen, Tübingen, Germany\\
\hline
\small David Mengel &  \small Department of Neurodegenerative Diseases, Hertie-Institute for Clinical Brain Research and Center of Neurology, University of Tübingen, Tübingen, Germany\\
\hline
\small Lisa Graf &  \small Department of Neurodegenerative Diseases, Hertie-Institute for Clinical Brain Research and Center of Neurology, University of Tübingen, Tübingen, Germany\\
\hline
\small Annick Vogels &  \small Department of Human Genetics, KU Leuven, Leuven, Belgium\\
\hline
\small Mathieu Vandenbulcke &  \small Geriatric Psychiatry Service, University Hospitals Leuven, Belgium; Neuropsychiatry, Department of Neurosciences, KU Leuven, Leuven, Belgium\\
\hline
\small Philip Van Damme &  \small Neurology Service, University Hospitals Leuven, Belgium; Laboratory for Neurobiology, VIB-KU Leuven Centre for Brain Research, Leuven, Belgium\\
\hline
\small Rose Bruffaerts &  \small Department of Biomedical Sciences, University of Antwerp, Antwerp, Belgium; Biomedical Research Institute, Hasselt University, 3500 Hasselt, Belgium\\
\hline
\small Pedro Rosa-Neto &  \small Translational Neuroimaging Laboratory, McGill Centre for Studies in Aging, McGill University, Montreal, Québec, Canada\\
\hline
\small Maxime Montembault &  \small Douglas Research Centre, Department of Psychiatry, McGill University, Montreal, Québec, Canada\\
\hline
\small Raphaella (Lara) Migliaccio &  \small Sorbonne Université, Paris Brain Institute – Institut du Cerveau – ICM, Inserm U1127, CNRS UMR 7225, AP-HP - Hôpital Pitié-Salpêtrière, Paris, France\\
\hline
\small Ninon Burgos &  \small Sorbonne Université, Paris Brain Institute – Institut du Cerveau – ICM, Inserm U1127, CNRS UMR 7225, AP-HP - Hôpital Pitié-Salpêtrière, Paris, France\\
\hline
\small Daisy Rinaldi &  \small Sorbonne Université, Paris Brain Institute – Institut du Cerveau – ICM, Inserm U1127, CNRS UMR 7225, AP-HP - Hôpital Pitié-Salpêtrière, Paris, France\\
\hline
\small Catharina Prix &  \small Neurologische Klinik, Ludwig-Maximilians-Universität München, Munich, Germany\\
\hline
\small Elisabeth Wlasich &  \small Neurologische Klinik, Ludwig-Maximilians-Universität München, Munich, Germany\\
\hline
\small Olivia Wagemann &  \small Neurologische Klinik, Ludwig-Maximilians-Universität München, Munich, Germany\\
\hline
\small Sonja Schönecker &  \small Neurologische Klinik, Ludwig-Maximilians-Universität München, Munich, Germany\\
\hline
\small Alexander Maximilian Bernhardt &  \small Neurologische Klinik, Ludwig-Maximilians-Universität München, Munich, Germany\\
\hline
\small Anna Stockbauer &  \small Neurologische Klinik, Ludwig-Maximilians-Universität München, Munich, Germany\\
\hline
\small Jolina Lombardi &  \small Department of Neurology, University of Ulm, Ulm\\
\hline
\small Sarah Anderl-Straub &  \small Department of Neurology, University of Ulm, Ulm, Germany\\
\hline
\small Adeline Rollin &  \small CHU, CNR-MAJ, Labex Distalz, LiCEND Lille, France\\
\hline
\small Gregory Kuchcinski &  \small Univ Lille, France; Inserm 1172, Lille, France; CHU, CNR-MAJ, Labex Distalz, LiCEND Lille, France\\
\hline
\small Maxime Bertoux &  \small Inserm 1172, Lille, France; CHU, CNR-MAJ, Labex Distalz, LiCEND Lille, France\\
\hline
\small Thibaud Lebouvier &  \small Univ Lille, France; Inserm 1172, Lille, France; CHU, CNR-MAJ, Labex Distalz, LiCEND Lille, France\\
\hline
\small Vincent Deramecourt &  \small Univ Lille, France; Inserm 1172, Lille, France; CHU, CNR-MAJ, Labex Distalz, LiCEND Lille, France\\
\hline
\small João Durães &  \small Neurology Department, Centro Hospitalar e Universitario de Coimbra, Coimbra, Portugal\\
\hline
\small Marisa Lima &  \small Neurology Department, Centro Hospitalar e Universitario de Coimbra, Coimbra, Portugal\\
\hline
\small Maria João Leitão &  \small Centre of Neurosciences and Cell Biology, Universidade de Coimbra, Coimbra, Portugal\\
\hline
\small Maria Rosario Almeida &  \small Faculty of Medicine, University of Coimbra, Coimbra, Portugal\\
\hline
\small Miguel Tábuas-Pereira &  \small Neurology Department, Centro Hospitalar e Universitario de Coimbra, Coimbra, Portugal; Faculty of Medicine, University of Coimbra, Coimbra, Portugal\\
\hline
\small Sónia Afonso &  \small Instituto Ciencias Nucleares Aplicadas a Saude, Universidade de Coimbra, Coimbra, Portugal\\
\hline
\small João Lemos &  \small Faculty of Medicine, University of Coimbra, Coimbra, Portugal\\
\bottomrule
\end{longtable}

\newpage

\paragraph*{Fig. S1}
\label{sfig:all_factors}
\textbf{Robust factors identified by sparse GFA.} \textbf{(a)} Total covariance explained by all robust factors. \textbf{(b)} Factor contributions for each genetic group. \textbf{(c)} Latent variables inferred by each factor grouped by genetic group. 
\begin{figure}[H]
\centering
\includegraphics[width=\linewidth]{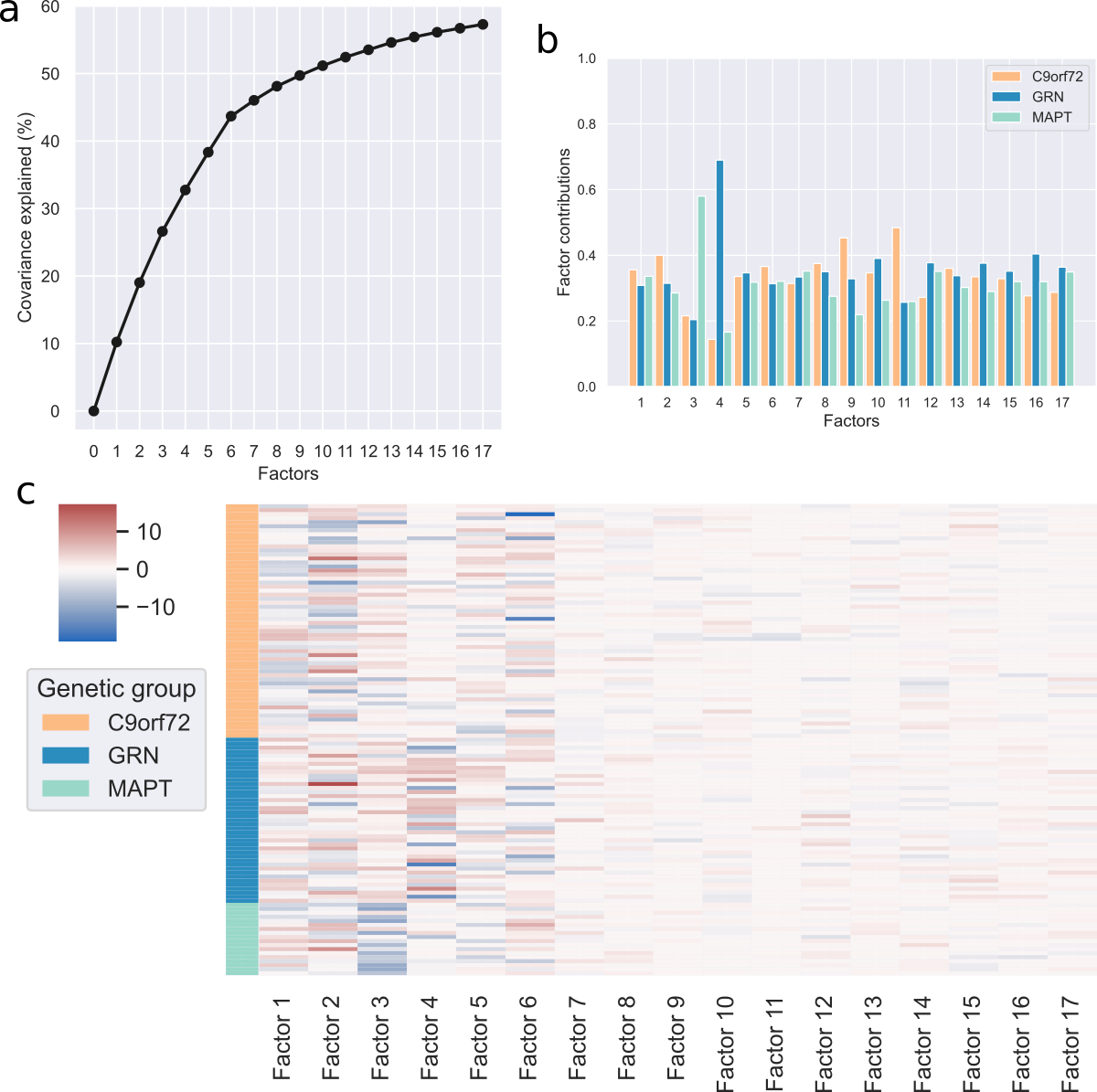}       
\end{figure}

\newpage

\paragraph*{Fig. S2}
\label{sfig:data_comps}
{\bf Top 4 factors identified by sparse GFA projected onto the data space.} \textbf{(a)} First, \textbf{(b)} second, \textbf{(c)} third and \textbf{(d)} fourth factors projected onto the data space. Blue represents a value below the sample average and red represents a value above the sample average.
\begin{figure}[H]
\centering
\includegraphics[width=\linewidth]{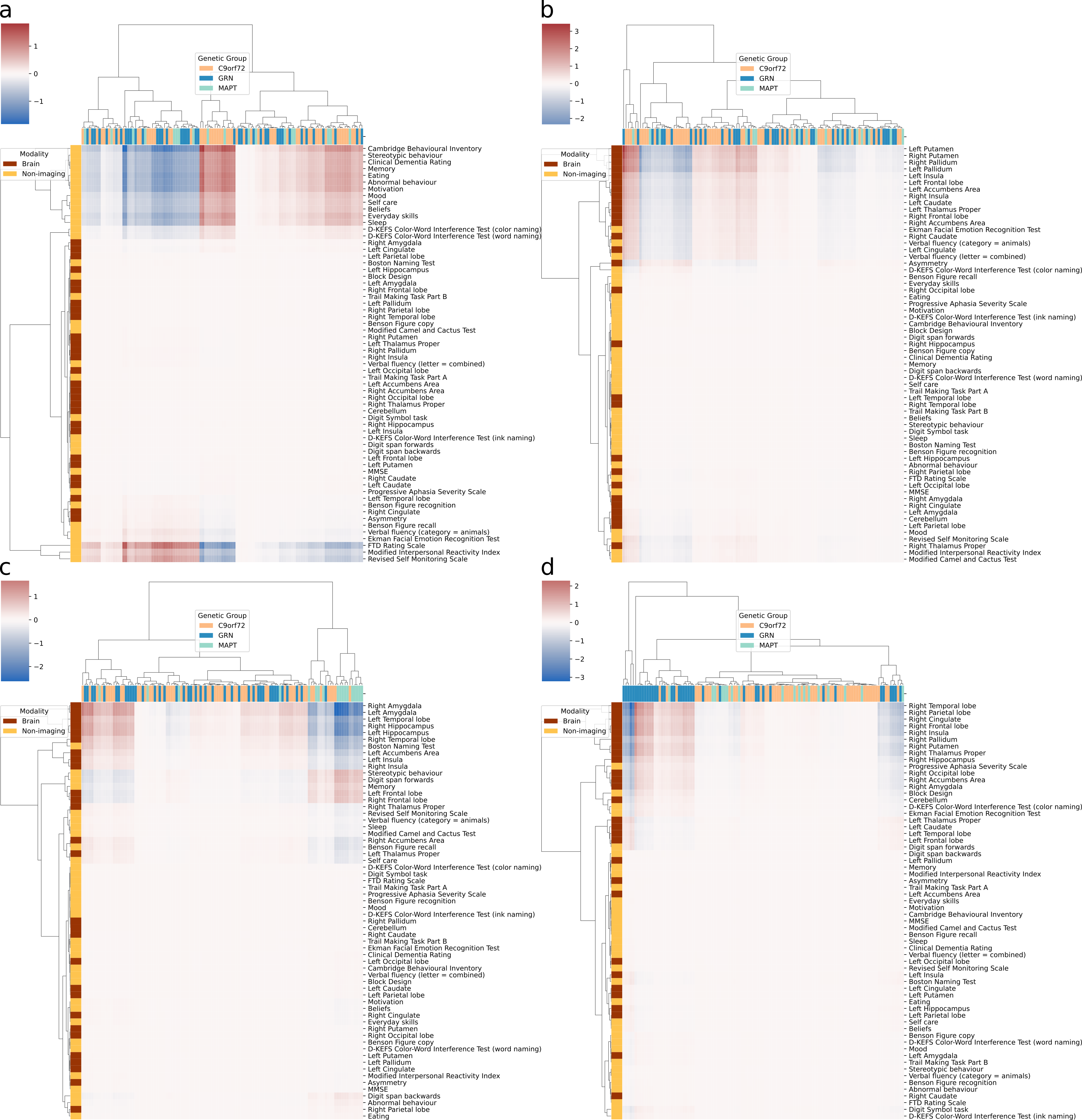}       
\end{figure}

\newpage

\paragraph*{Fig. S3}
\label{sfig:all_brain_loadings}
{\bf Brain loadings of all robust factors identified by sparse GFA.}
\begin{figure}[H]
\centering
\includegraphics[width=0.78\linewidth]{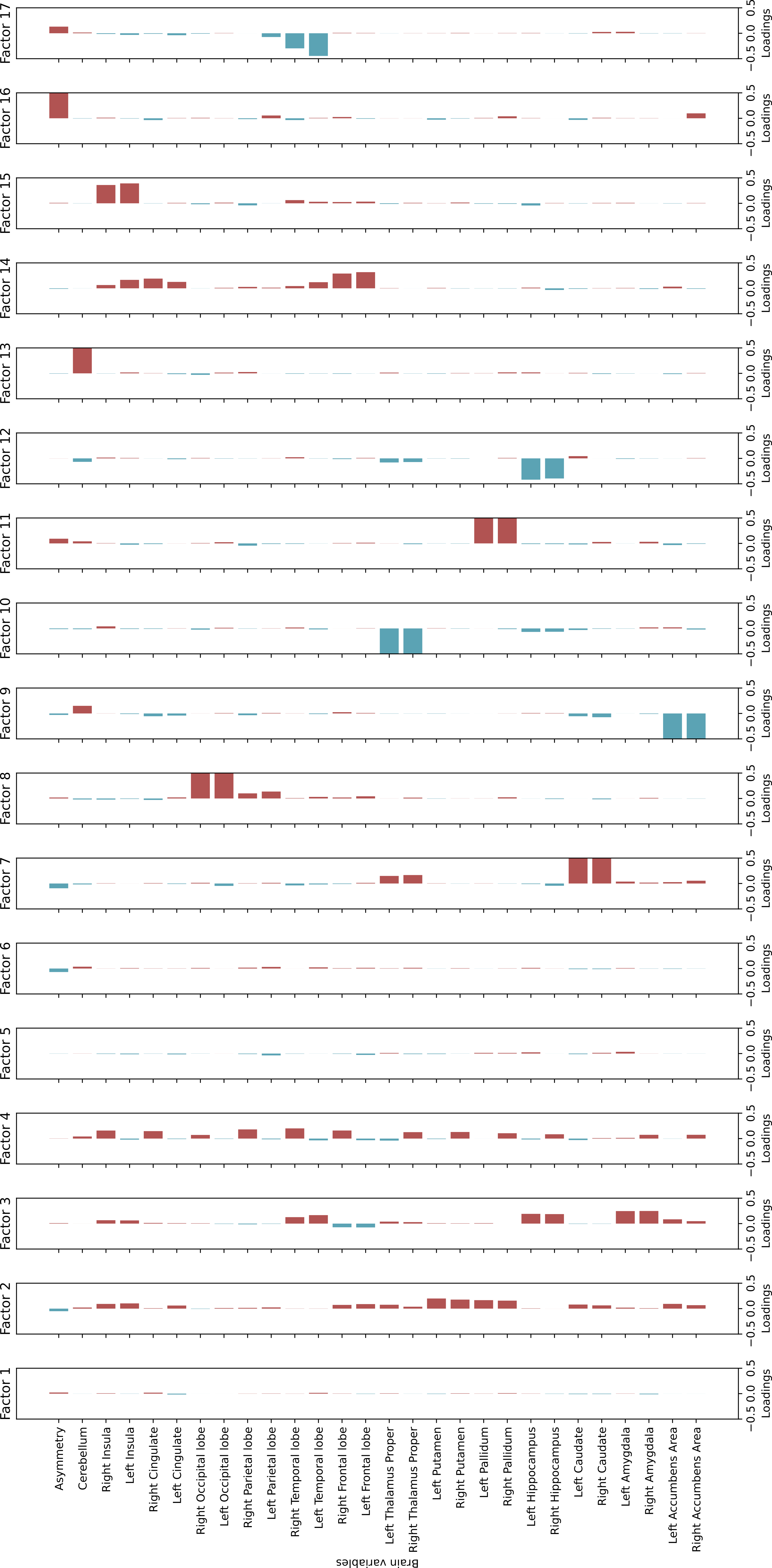}       
\end{figure}

\newpage

\paragraph*{Fig. S4}
\label{sfig:all_NI_loadings}
{\bf Non-imaging loadings of all robust factors identified by sparse GFA.}
\begin{figure}[H]
\centering
\includegraphics[width=0.60\linewidth]{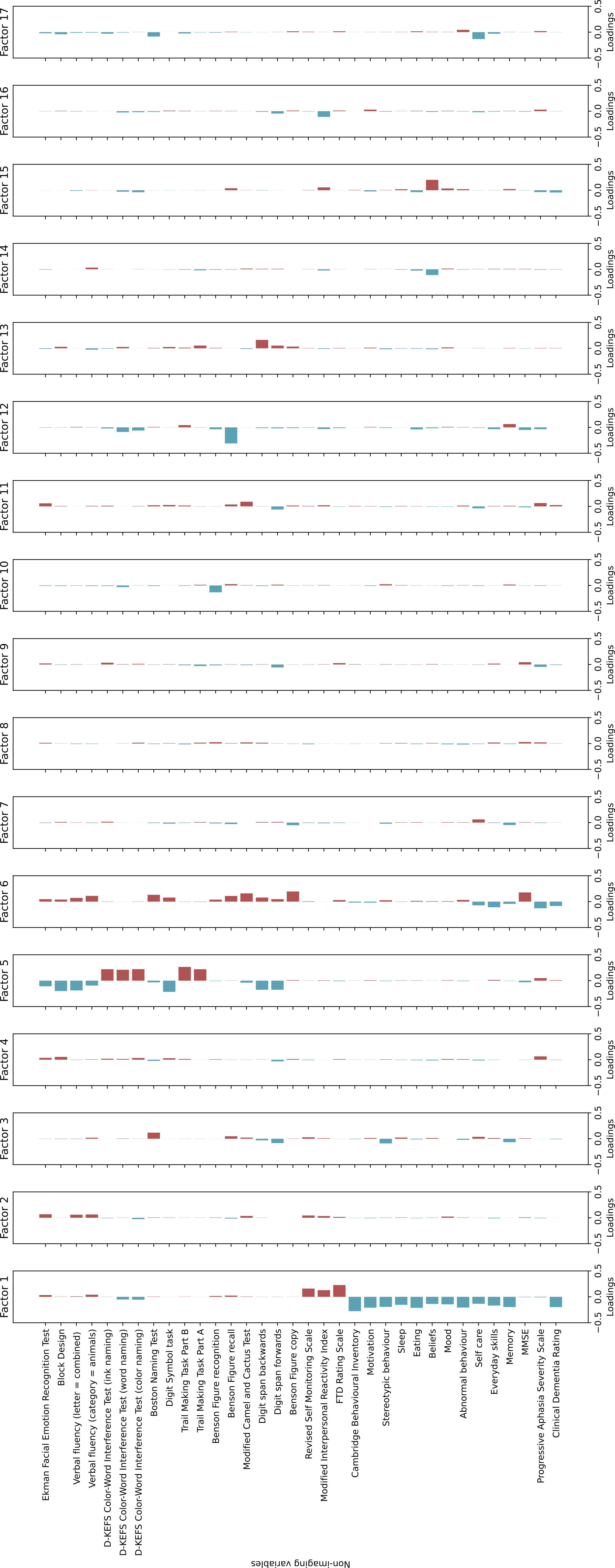}       
\end{figure}

\newpage

\paragraph*{Table S1}
\label{stab:NI_vars}
{\bf Description of the non-imaging features of the GENFI dataset.} Orange corresponds to medical assessments of the patients; blue corresponds to informant questionnaires completed by the primary caregiver; red corresponds to neuropsychological tasks completed by patients. $^{*}$a high score in a test means that the individual is more affected; $^{**}$a high score in a test means that the individual is less affected; $^{+}$subscore of the Cambridge Behavioural Inventory.
\begin{table}[H]
    \begin{center}
      \begin{tabular}{p{0.45\textwidth}|p{0.4\textwidth}}
        \toprule 
        \textbf{Variable name} & \textbf{Cognitive function}\\
        \midrule 
        \textcolor{orange}{Clinical Dementia Rating$^{*}$} & \textcolor{orange}{disease severity (FTD appropriate)}\\
        \textcolor{orange}{Progressive Aphasia Severity Scale$^{*}$} & \textcolor{orange}{speech and language}\\
        \textcolor{orange}{Mini Mental State Examination$^{**}$} & \textcolor{orange}{disease severity (general dementia)}\\
        \textcolor{blue}{Memory$^{*,+}$} & \textcolor{blue}{memory}\\
        \textcolor{blue}{Everyday skills$^{*,+}$} & \textcolor{blue}{everyday skills}\\
        \textcolor{blue}{Self care$^{*,+}$} & \textcolor{blue}{self care}\\
        \textcolor{blue}{Abnormal behaviour$^{*,+}$} & \textcolor{blue}{abnormal behaviour}\\
        \textcolor{blue}{Mood$^{*,+}$} & \textcolor{blue}{mood}\\
        \textcolor{blue}{Beliefs$^{*,+}$} & \textcolor{blue}{beliefs}\\
        \textcolor{blue}{Eating$^{*,+}$} & \textcolor{blue}{eating}\\
        \textcolor{blue}{Sleep$^{*,+}$} & \textcolor{blue}{sleep}\\
        \textcolor{blue}{Stereotypic behaviour$^{*,+}$} & \textcolor{blue}{stereotypic behaviour}\\
        \textcolor{blue}{Motivation$^{*,+}$} & \textcolor{blue}{motivation}\\
        \textcolor{blue}{Cambridge Behavioural Inventory$^{*}$} & \textcolor{blue}{general behaviour}\\
        \textcolor{blue}{FTD Rating Scale$^{**}$} & \textcolor{blue}{disease severity}\\
        \textcolor{blue}{Modified Interpersonal Reactivity Index$^{**}$} & \textcolor{blue}{cognitive and emotional empathy}\\
        \textcolor{blue}{Revised Self Monitoring Scale$^{**}$} & \textcolor{blue}{social behaviour}\\
        \textcolor{red}{Benson Figure copy$^{**}$} & \textcolor{red}{visuospatial skills}\\
        \textcolor{red}{Benson Figure recall$^{**}$} & \textcolor{red}{episodic memory}\\
        \textcolor{red}{Benson Figure recognition$^{**}$} & \textcolor{red}{visual memory}\\
        \textcolor{red}{Digit span forwards$^{**}$} & \textcolor{red}{attention processing}\\
        \textcolor{red}{Digit span backwards$^{**}$} & \textcolor{red}{working memory}\\
        \textcolor{red}{Modified Camel and Cactus Test$^{**}$} & \textcolor{red}{semantic memory}\\
        \textcolor{red}{Trail Making Test Part A$^{*}$} & \textcolor{red}{attention processing}\\
        \textcolor{red}{Trail Making Test Part B$^{*}$} & \textcolor{red}{executive function}\\
        \textcolor{red}{Digit Symbol task$^{**}$} & \textcolor{red}{processing speed}\\
        \textcolor{red}{Boston Naming Test$^{**}$} & \textcolor{red}{object naming - word retrieval}\\
        \textcolor{red}{D-KEFS Color-Word Interference Test (word naming)$^{*}$} & \textcolor{red}{executive function}\\
        \textcolor{red}{D-KEFS Color-Word Interference Test (color naming)$^{*}$} & \textcolor{red}{executive function}\\
        \textcolor{red}{D-KEFS Color-Word Interference Test (ink naming)$^{*}$} & \textcolor{red}{executive function}\\
        \textcolor{red}{Verbal fluency (category = animals)$^{**}$} & \textcolor{red}{language fluency}\\
        \textcolor{red}{Verbal fluency (letter = combined)$^{**}$} & \textcolor{red}{language fluency}\\
        \textcolor{red}{Block Design$^{**}$} & \textcolor{red}{visuospatial skills}\\
        \textcolor{red}{Ekman Facial Emotion Recognition Test$^{**}$} & \textcolor{red}{emotion recognition}\\
        \bottomrule 
      \end{tabular}
    \end{center}
\end{table}

\newpage

\bibliographystyle{apalike-oadoi}
\bibliography{main}
\end{document}